\documentclass{article}

\usepackage{microtype}
\usepackage{graphicx}
\usepackage{subfigure}
\usepackage{booktabs} % for professional tables
\usepackage{makecell}
\usepackage{listings}
\usepackage{amsmath}

\usepackage{color, colortbl}
\usepackage{xcolor}
\definecolor{greyC}{RGB}{180,180,180}
\definecolor{greyL}{RGB}{235,235,235}

\definecolor{codegreen}{rgb}{0,0.6,0}
\definecolor{codegray}{rgb}{0.5,0.5,0.5}
\definecolor{codepurple}{rgb}{0.58,0,0.82}
\definecolor{backcolour}{rgb}{0.95,0.95,0.92}

\lstdefinestyle{mystyle}{
    backgroundcolor=\color{backcolour},   
    commentstyle=\color{codegreen},
    keywordstyle=\color{magenta},
    numberstyle=\tiny\color{codegray},
    stringstyle=\color{codepurple},
    basicstyle=\ttfamily\footnotesize,
    breakatwhitespace=false,         
    breaklines=true,                 
    captionpos=b,                    
    keepspaces=true,                 
    numbers=left,                    
    numbersep=5pt,                  
    showspaces=false,                
    showstringspaces=false,
    showtabs=false,                  
    tabsize=2
}
\usepackage{enumitem}
\setlist[itemize]{leftmargin=*,align=parleft,left=0pt..1em}

\usepackage{hyperref}
\usepackage[hyphenbreaks]{breakurl}
\usepackage{url}

\usepackage{amssymb}

\usepackage[accepted]{mlsys2022}

\mlsystitlerunning{FedML Parrot}

\begin{document}

\twocolumn[
% \mlsystitle{Parrot: Scalable Federated Learning System via Heterogeneity-aware Scheduling on \\ Sequential and Hierarchical Training}
\mlsystitle{FedML Parrot: A Scalable Federated Learning System via Heterogeneity-aware Scheduling on Sequential and Hierarchical Training}

% unless you've provided the [accepted] option to the mlsys2022


% \mlsyssetsymbol{equal}{*}

\begin{mlsysauthorlist}
\mlsysauthor{Zhenheng Tang}{hkbu}
\mlsysauthor{Xiaowen Chu}{hkbu}
\mlsysauthor{Ryan Yide Ran}{fedml}
\mlsysauthor{Sunwoo Lee}{Inha}
\mlsysauthor{Shaohuai Shi}{hkust}
\mlsysauthor{Yonggang Zhang}{hkbu}
\mlsysauthor{Yuxin Wang}{hkbu}
\mlsysauthor{Alex Qiaozhong Liang}{fedml}
\mlsysauthor{Salman Avestimehr}{fedml}
\mlsysauthor{Chaoyang He}{fedml}
\end{mlsysauthorlist}

\mlsysaffiliation{hkbu}{Hong Kong Baptist University}
\mlsysaffiliation{fedml}{FedML Inc.}
\mlsysaffiliation{Inha}{Inha University}
\mlsysaffiliation{hkust}{The Hong Kong University of Science and Technology}

% \mlsysaffiliation{hkustgz}{The Hong Kong University of Science and Technology (Guangzhou)}

\mlsyscorrespondingauthor{Chaoyang He}{ch@fedml.ai}

\mlsyskeywords{Machine Learning, MLSys}

\vskip 0.3in

\begin{abstract}
Federated Learning (FL) enables collaborations among clients for train machine learning models while protecting their data privacy. Existing FL simulation platforms that are designed from the perspectives of traditional distributed training, suffer from laborious code migration between simulation and production, low efficiency, low GPU utility, low scalability with high hardware requirements and difficulty of simulating stateful clients. In this work, we firstly demystify the challenges and bottlenecks of simulating FL, and design a new FL system named as FedML \texttt{Parrot}. It improves the training efficiency, remarkably relaxes the requirements on the hardware, and supports efficient large-scale FL experiments with stateful clients by: (1) sequential training clients on devices; (2) decomposing original aggregation into local and global aggregation on devices and server respectively; (3) scheduling tasks to mitigate straggler problems and enhance computing utility; (4) distributed client state manager to support various FL algorithms.  Besides, built upon our generic APIs and communication interfaces, users can seamlessly transform the simulation into the real-world deployment without modifying codes. We evaluate \texttt{Parrot} through extensive experiments for training diverse models on various FL datasets to demonstrate that \texttt{Parrot} can achieve simulating over 1000 clients (stateful or stateless) with flexible GPU devices setting ($4 \sim 32$) and high GPU utility, 1.2 $\sim$ 4 times faster than FedScale, and 10 $\sim$ 100 times memory saving than FedML. And we verify that \texttt{Parrot} works well with homogeneous and heterogeneous devices in three different clusters. Two FL algorithms with stateful clients and four algorithms with stateless clients are simulated to verify the wide adaptability of \texttt{Parrot} to different algorithms. Code will be merged into \url{https://github.com/FedML-AI/FedML}.

\end{abstract}
]



\printAffiliationsAndNotice{}  % leave blank if no need to mention equal contribution
% \printAffiliationsAndNotice{\mlsysEqualContribution} % otherwise use the standard text.

\section{Introduction}
Federated Learning (FL) is a distributed machine learning paradigm enabling collaborative training between multiple clients with data privacy protection~\citep{mcmahan2017communication,kairouz2019advances}. It supports organizations and personal users who are sensitive to privacy issues to learn powerful models without sharing data with each other. Many applications have been benefited from FL with a large number of clients, like word prediction~\citep{hard2018federated}, object detection~\citep{liu2020fedvision,hsu2020federated} and healthcare~\citep{fedhealth,FLCOVID}.
Different from distributed training~\citep{li2014scaling} in a data center, the number of clients in FL varies from $ 2\sim 10^{10}$~\citep{hard2018federated,kairouz2019advances}, making the scaling of federated learning extremely challenging.

One important research area that is largely ignored by the current FL community is as follows. Before real-world deployment, new FL models and algorithms usually need to be verified by simulation. But, how can we enable a rapid and flexible prototyping for simulating an arbitrary number of clients (i.e., large-scale) in a fixed small-scale GPU cluster, while simultaneously guaranteeing zero-code change towards production deployment without laborious and erroneous code migration?

\begin{figure*}[htb!] 
\small
   \centering
    \includegraphics[width=1.0\linewidth]{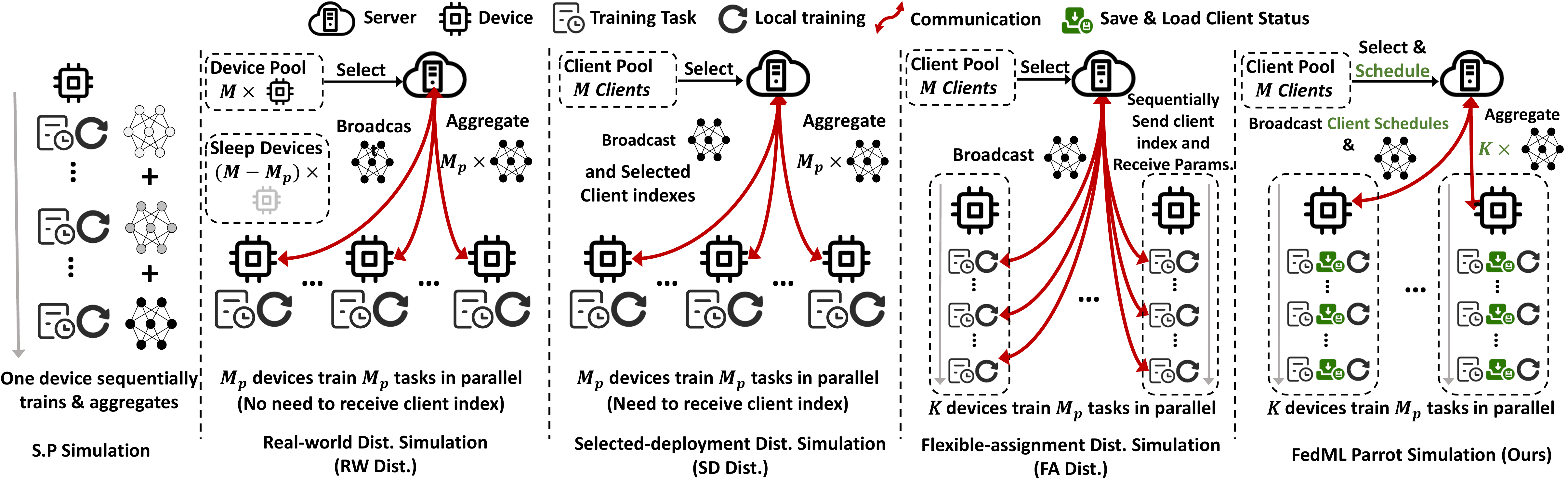} 
    \vspace{-0.5cm}
    \caption{Overview of different FL simulation systems.}
    \label{fig:FLsimulation} 
\vspace{-0.5cm}
\end{figure*}
There exist many FL frameworks supporting different FL algorithms running on GPU/CPU computing resource~\citep{caldas2018leaf,crosssilononiid,fedscale,beutel2020flower}. However, these frameworks are far from meeting the research exploration demands mentioned above. They either require the number of devices to be the same as the number of clients, or they only use a single device to simulate all clients~\citep{caldas2018leaf,crosssilononiid}. Both approaches cannot support large-scale training of FL algorithms. To alleviate the high requirement of hardware, some FL frameworks~\citep{fedscale,beutel2020flower} enable multiple clients to run on a single executor at each training round and the executor may train the models from different clients at different rounds, like the simulation using docker~\citep{8113094}. Yet, there still exist a few main problems in these systems. 

First, the communication overhead between clients and the server is very significant making the computing resources highly under-utilized. Second, clients typically have very high data heterogeneity (e.g., different sizes), so the training time of different clients is quite different. This is a classical straggler problem, where the server must wait for the slowest executor to finish its tasks, causing long training time and very low GPU/TPU utilization. 

Third, these frameworks ignore the stateful clients in some FL algorithms, where executors need to store the historical state of clients so that each executor can run different clients at different rounds. Many FL variant algorithms~\citep{fedprox,karimireddy2019scaffold,wang2020tackling,acar2021federated,luo2021no,li2021model,chen2021bridging,exploitsharedRep} are developed to tackle the data heterogeneity problem where clients typically have different data distributions and/or various data sizes, making simple FL algorithms, like FedAvg, difficult to converge and leads to bad generalization performance~\citep{woodworth2020minibatch,acar2021federated}. These algorithms may not be limited to exchanging model parameters during training, but possibly include other parameters like intermediate features~\citep{exploitsharedRep}, masks of model~\citep{fedmask}, auxiliary gradient corrections~\citep{karimireddy2019scaffold}, third-party datasets~\citep{FedDF,VHL}, etc. Moreover, many FL algorithms require stateful clients to store some client state, like the control variates~\citep{karimireddy2019scaffold}, old gradients~\citep{acar2021federated}, personalized models or layers~\citep{liang2020think,chen2021bridging}, model masks~\citep{fedmask} etc. 

Fourth, in the current community, the cost of migration is high and requires collaboration from both research team and product team; the simulation is unable to quickly find problems in the deployment; A small error in the code inconsistency will lead to errors in the training algorithm, and eventually a failure of training effective model.

To support large-scale FL experiments with flexible extension and design of algorithms, a general FL simulation system is highly demanded. Aiming to address the above problems, we make the following contributions in this work.

\textbf{Contributions.} We first systemically demystify the challenges and bottlenecks of simulating FL algorithms (\S\ref{sec:demystify}). Then, we propose a new efficient and flexible FL system named \texttt{Parrot} (\S\ref{sec:systemdesign}\S\ref{sec:perfOpt}) with the following four features. 1) We design a hierarchical data aggregation mechanism, where the client results are first aggregated locally on the devices, and then aggregated globally at the server, to significantly reduce the communication complexity. 2) We design a task scheduler with estimating the dynamic real-time workloads to minimize the training time of one round, which alleviates the straggler problem and improves the GPU utilization. 3) We design a client state manager to support stateful-client algorithms. 4) \texttt{Parrot} provides generic APIs and abstract communication layers so that users can easily extend their new algorithms in the simulation. The well-verified algorithms and models in the simulation can be seamlessly deployed in real-world FL environments.

We verify the flexibility and efficiency of \texttt{Parrot}, through extensive experiments, including training diverse models (ResNet18, ResNet50~\citep{he2016deep}, and Albert~\citep{Lan2020ALBERT}) on various datasets (FEMNIST~\citep{caldas2018leaf}, ImageNet~\citep{russakovsky2015imagenet}, and Reddit~\citep{fedscale}) with different number of devices and clients (i.e., 1000 and 10000 clients). Experiments in three different clusters with homogeneous and heterogeneous devices verify that \texttt{Parrot} works well with different hardware environments. Specifically, \texttt{Parrot} successfully simulates two FL algorithms with stateful clients and four algorithms with stateless clients showing the high flexibility of \texttt{Parrot} in supporting different FL algorithms. Experimental results also show that \texttt{Parrot} achieves a nearly linear speedup with an increased number of devices on different hardware configurations. 

% First, \texttt{Parrot} assigns devices with multiple client simulation tasks, enabling simulating extremely large number of concurrent clients with flexible GPU devices setting. Second, client results are aggregated locally on the devices, and aggregated globally at the server, reducing the communication complexity. Third, a task scheduler is designed to estimate the dynamic real-time workloads and minimize the training time of one round, alleviating the straggler problem and enhancing the GPU utility. Last, a client state manager is designed to support stateful-client algorithms.

% $\colorbox{blue!20}{$s_d$}$)M)$
\begin{table*}[h!]
\caption{Complexity comparison between different simulation schemes. Note \textbf{client state manager} can be used not only in \texttt{Parrot}, but also other schemes. $M$ represents the number of total clients, $M_p$ the number of selected clients per round. In \texttt{RW Dist.} and \texttt{SD Dist.}, the number of devices that join the training per round is $M_p$, and $K$ in \texttt{FA Dist.} and FedML \texttt{Parrot} ($K \le M_p$). $s_m$ represents the size of memory that is needed to simulate one client. $s_a$ represents the size AVG. parameters that are averaged on the server. ${\color{red}s_e}$ is the size of special parameters that should be collected by the server instead of averaging, which is only needed in a part of algorithms. ${\color{blue}s_d}$ is the size of client state, which is only needed by algorithms with stateful clients.}
\vspace{-0.3cm}
\setlength\tabcolsep{3pt}
\label{tab:complexity}
% \vskip 0.15in
\begin{center}
% \begin{small}
% \begin{sc}
% \scriptsize{
% \small{
\resizebox{\textwidth}{!}{
\begin{tabular}{lccccc}
\toprule
Scheme & \texttt{SP} & \texttt{RW Dist.} & \texttt{SD Dist.} & \texttt{FA Dist.} & FedML \texttt{Parrot} \\
\bottomrule
Number of Devices (Executors) & 1  & $M$   &  $M_p$ & $K$ &  $K$  \\
Memory &  $O(s_mM + {\color{blue}{s_d}}M)$  &   $O(s_mM + {\color{blue}{s_d}}M)$ & $O(s_mM_p + {\color{blue}{s_d}}M)$ &  $O(s_mK + {\color{blue}{s_d}}M)$ &  $O(s_mK + {\color{blue}{s_d}}M)$ \\
Memory with \textbf{state manager}  & $O(s_m + {\color{blue}{s_d}})$  &  $O(s_mM + {\color{blue}{s_d}}M_p)$ & $O(s_mM_p + {\color{blue}{s_d}}M_p)$ &  $O(s_mK + {\color{blue}{s_d}}K)$ &  $O(s_mK + {\color{blue}{s_d}}K)$ \\
Disk Cost with \textbf{state manager}  & $O(s_mM_p+{\color{blue}{s_d}}M)$  &  $O({\color{blue}{s_d}}M)$ & $O({\color{blue}{s_d}}M)$ &  $O({\color{blue}{s_d}}M)$ &  $O({\color{blue}{s_d}}M)$ \\
Comm. Size  & $O(0)$ &   $O(s_aM_p + {\color{red}{s_e}}M_p)$ & $O(s_aM_p + {\color{red}{s_e}}M_p)$ & $O(s_aM_p + {\color{red}{s_e}}M_p)$ &  $O(s_aK + {\color{red}{s_e}}M_p)$  \\
Comm. Trips &  $O(0)$ & $O(M_p)$  & $O(M_p)$ & $O(M_p)$ & $O(K)$ \\
\bottomrule
\end{tabular}
% \vspace{0pt}
}
% \begin{tablenotes}
% 	\item  
% \end{tablenotes} 
% \end{sc}
% \end{small}
\end{center}
% \vskip -0.1in
% \vspace{0pt}
\vspace{-0.5cm}
\end{table*}

\section{Background and Challenges}\label{sec:demystify}
% \vspace{-0.3cm}
\subsection{Federated Learning}
% Suppose there are participating clients $\mathcal{M} = \left\{1,2,\cdots,M  \right\}$. FL aims collaborate clients to learn a global or personalized machine learning model $\theta $ with their own distribution $\mathcal{D}_m$. Formally, the global optimization problem of FL is commonly formulated as~\citep{mcmahan2017communication,fedprox}:
% \begin{equation}\label{eq:F}
%     \min_{\theta\in \mathbb{R}^d} F(\theta) := \sum_{m=1}^{M} p_m F_m(\theta)    = \sum_{m=1}^{M} p_m \mathbb{E}_{\xi_m \sim \mathcal{D}_m} f(\theta;\xi_m),
% \end{equation}
% where $F_m(\theta)=\mathbb{E}_{\xi_m \sim \mathcal{D}_m} f(\theta;\xi_m)$ is the local objective function, $p_m > 0 $ and $ \sum_{m=1}^{M} p_m =1  $. Usually, $p_m$ is set as $\frac{N_m}{N}$, where $N_m$ denotes the number of samples on client $m$ and $N=\sum_{m=1}^{M} N_m$. 

FedAvg is a very general learning algorithm in FL, which iteratively updates model parameters until convergence over many training rounds. In each round, a central server first randomly selects a part of clients (called concurrent clients), and broadcasts the global parameter to these clients. Then, the selected clients conduct model training with their private data and send the newly updated local parameters to the server. After that, the server aggregates local parameters to obtain new global parameters and then goes to the next training round.

\begin{figure}[ht!] 
\small
   \centering
    \includegraphics[width=0.90\linewidth]{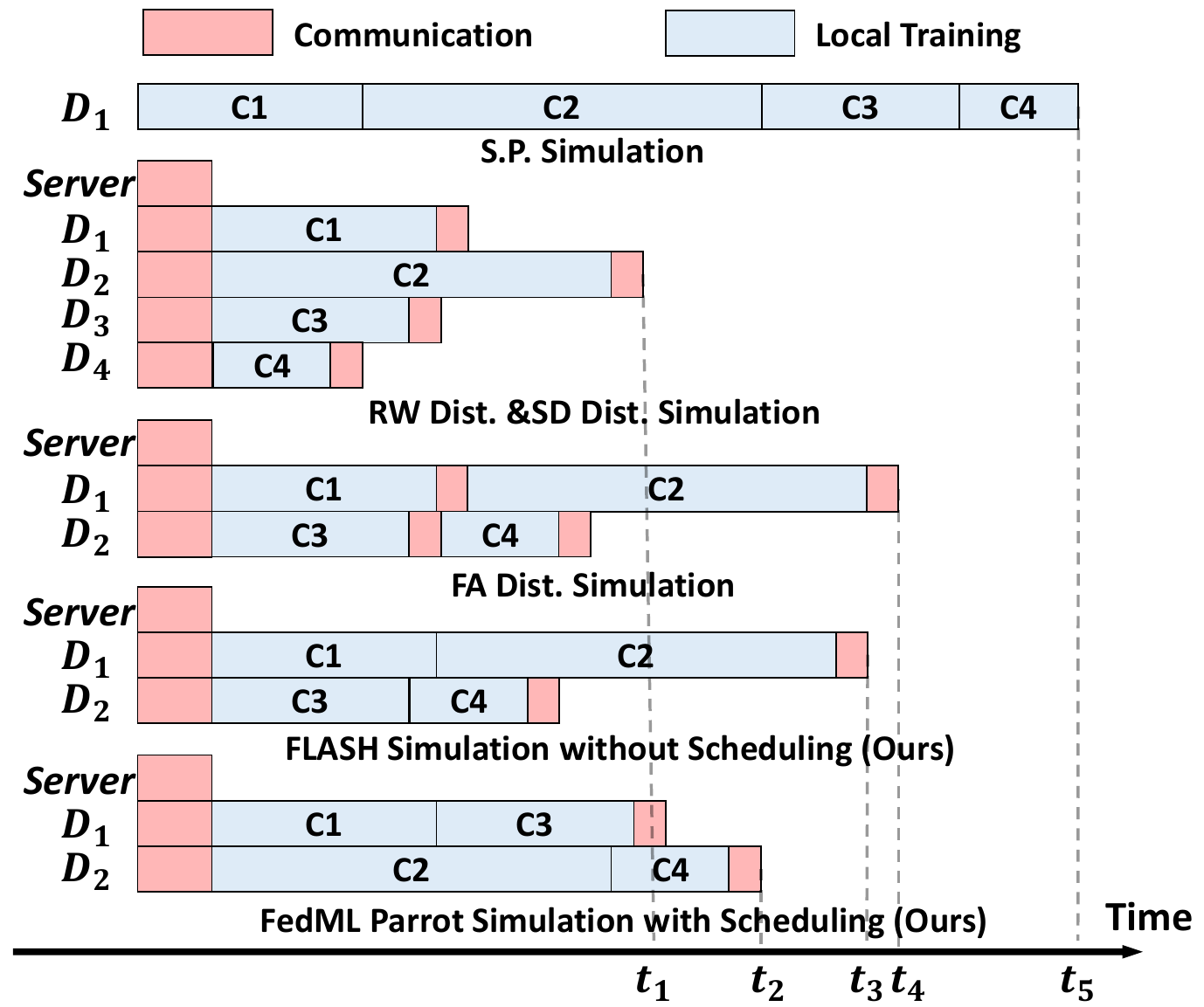} 
    \vspace{-0.3cm}
    \caption{Timelines of different simulation schemes, where $D_i$ represents the $i$-th device.} 
    \label{fig:FLtimelines} 
\vspace{-0.5cm}
\end{figure}
% \vspace{-0.1cm}

\subsection{FL Simulation Frameworks}
% \subsection{Demystifying current FL simulation schemes}
Existing state-of-the-art FL simulation frameworks~\citep{caldas2018leaf,TFF2019,chaoyanghe2020fedml,ryffel2018generic,fedscale,beutel2020flower} can be categorized into two types: single-process simulation and distributed simulation. 
% We demystify their differences in Figure~\ref{fig:FLsimulation}. And the timeline of training one FL round is shown in Figure~\ref{fig:FLtimelines}.

\begin{table*}[ht!]
\caption{Comparisons between different FL frameworks.}
\label{sample-table}
\begin{center}
\begin{small}
\begin{tabular}{ccccccccc}
\toprule
 &  LEAF & TFF & Syft & FedScale &  Flower & FederatedScope &  FedML Cross-silo  & FedML Parrot \\
\bottomrule
\texttt{SP} & $\checkmark$  & $\checkmark$&$\checkmark$  &$\checkmark$ &$\checkmark$ &$\checkmark$ &$\checkmark$  & $\checkmark$  \\
\texttt{RW Dist.} & &  &  &  $\checkmark$ &  & $\checkmark$ & $\checkmark$ & $\checkmark$  \\
\texttt{SD Dist.} & & &  & & $\checkmark$ &  & $\checkmark$& $\checkmark$  \\
\texttt{FA Dist.} & & &  & $\checkmark$ & $\checkmark$ &  & & $\checkmark$  \\
Scalability & & $\checkmark$ &  & $\checkmark$  & $\checkmark$ & $\checkmark$ & $\checkmark$ & $\checkmark$  \\
Flexible Hardware Conf. &   $\checkmark$ & &  & $\checkmark$  & $\checkmark$ & &  & $\checkmark$  \\
Real-world Deployment & & &  & $\checkmark$ & $\checkmark$ & $\checkmark$ & $\checkmark$  & $\checkmark$  \\
Task Scheduling  & & &  & & &  &  & $\checkmark$ \\
Client State Manager  & & &  & & &  &  & $\checkmark$ \\
% \midrule
\bottomrule
\end{tabular}
\end{small}
\end{center}
\vskip -0.1in
\label{tab:related}
\end{table*}

\textbf{Single-process Simulation} (\texttt{SP}).
The training tasks of different clients are sequentially run on a computational device in \texttt{SP}. Representative systems include LEAF~\citep{caldas2018leaf} and TFF~\citep{TFF2019}. Since all models are stored on a single device, \texttt{SP} can avoid communications in aggregating models as shown in Figure~\ref{fig:FLtimelines}. However, FL generally requires to simulate many clients with heterogeneous datasets, so a single device in \texttt{SP} should interchangeably train models for different clients, which is very time-consuming.

\textbf{Distributed Simulation}. Distributed simulation can be further classified into real-world distributed simulation (\texttt{RW Dist.}), selected-deployment distributed simulation (\texttt{SD Dist.}), and flexible-assignment distributed simulation (\texttt{FA Dist.}). 

% In \texttt{RW Dist.}, all clients (say $M$ clients) should be initialized on $M$ devices. It follows the FedAvg protocol shown in Figure~\ref{fig:FLoverview} during the training process. Representative systems include Syft~\citep{ryffel2018generic} and FederatedScope~\citep{federatedscope}). In each communication round, $M_p$ selected clients or devices for training, leaving the other $M - M_p$ un-selected devices idle. This dramatically under-utilizes computing resources and it becomes extremely costly when scaling to massive clients, e.g., $1000$ or $10000$ clients. 

In \texttt{RW Dist.}, all clients (say $M$ clients) should be initialized on $M$ devices. It follows the FedAvg protocol during the training process. Representative systems include Syft~\citep{ryffel2018generic} and FederatedScope~\citep{federatedscope}). In each communication round, $M_p$ selected clients or devices for training, leaving the other $M - M_p$ un-selected devices idle. This dramatically under-utilizes computing resources and it becomes extremely costly when scaling to massive clients, e.g., $1000$ or $10000$ clients.

In \texttt{SD Dist.}, it only requires the number of devices to be the same as the number of selected clients per round, i.e., $M_p$. In each round, each device receives the client index that is assigned by the server. Then devices will simulate the assigned client. This avoids the extremely low CPU or GPU utility caused by the idle devices. If the devices are homogeneous, \texttt{RW Dist.} and \texttt{SD Dist.} have the same timeline as shown in Figure~\ref{fig:FLtimelines} and their training time is the shortest among all schemes under the same environment. However, due to the data heterogeneity in different clients, \texttt{SD Dist.} and \texttt{RW Dist.} have the straggler problem. 

% More fatally, for many stateful algorithms, due to the partial joining property of FL, devices must know the state of the newly assigned clients in the current round. It is unrealistic to support each device to store all historical client state and communicate them to other clients in each round, which requires enormous extra memory and communication overhead.

% \textbf{Flexible-assignment Distributed Simulation} (\texttt{FA Dist.}).
\texttt{FA Dist.} can be seen as a combination of \texttt{SP} and distributed simulation. Representive systems include FedScale~\citep{fedscale} and Flower~\citep{beutel2020flower}. \texttt{FA Dist.} supports multiple devices for simulating and allows the number of devices, $K$, can be smaller than the number of selected clients, $M_p$. Each device sequentially simulates multiple clients. After simulating one client, each device instantly returns back the result to the server. Then, the server assigns a new client index that needs to be simulated in the current round to this device. \texttt{FA Dist.} significantly relaxes the hardware requirements of FL simulation. A few CPUs or GPUs can simulate large-scale FL. However, the straggler, large communication overhead and the stateful clients problems still exist as shown in Figure~\ref{fig:FLtimelines}.

\textbf{From Simulation to Production Deployment}.
Many FL frameworks are designed to simulate FL in the centralized  computing cluster~\citep{caldas2018leaf,TFF2019,fedscale,beutel2020flower}. The low-level communication APIs are based on MPI~\citep{MPI} or PyTorch-DDP~\citep{paszke2019pytorch}. It is hard to deploy the FL algorithms implemented in the simulation to the real-world deployment, which may need laborious code migration.

We summarize the system overheads of simulating one FL round of different schemes in Table~\ref{tab:complexity} with respect to the number of devices, CPU or GPU memory, disk storage, communication size, and communication trips. Note that the client state manager in our framework can be an auxiliary component to enhance other schemes. The communication trip means the number of communication rounds between devices and the server, which is an important factor that affects the communication time~\citep{CommModel,MGWFBP,papaya}. We provide detailed complexity analysis in Section~\ref{sec:complexity}.

\subsection{Challenges}

% Therefore, an efficient and flexible FL simulator is highly demanded by satisfying the following requirements. 
%     (1) Users who want to quickly can easily verify their FL algorithms or models.
%     (2) Users can design novel FL algorithms that need to communicate special parameters and/or have stateful clients.
%     (3) Users who have limited CPU or GPU devices are able to conduct large-scale FL simulation with a large number of concurrent clients and total clients.
%     (4) Users can \textit{conveniently deploy} their FL algorithms or models into real-world applications after successful experimentation on the simulator.

\texttt{Parrot} aims to offer users an efficient FL system to (1) accelerate the simulation remarkably; (2) reduce the requirements of a large number of edge devices for simulating large-scale FL experiments, e.g. 1000 $\sim$ 10000 clients; (3) support a large range of FL algorithms, including stateful federated optimizers; (4) provide seamless migration from the research simulation to real-world FL deployment, to address four following challenges in a unified framework.

\begin{itemize}[noitemsep,nolistsep]
\setlength{\itemsep}{0pt}
\setlength{\parsep}{0pt}
\setlength{\topsep}{0pt}
\setlength{\partopsep}{0pt}
\setlength{\parskip}{0pt}
    \item \textbf{Heterogeneous data}. Clients have different data distributions and dataset sizes~\citep{kairouz2019advances}. This leads to the straggler problem, i.e. the server must wait for the slowest clients. This also appears in the traditional distributed training~\citep{StaleASGD} but only when devices are heterogeneous.
    \item \textbf{Simulating massive clients on small cluster}. The number of clients in FL varies from 2 to over $10^{10}$~\citep{kairouz2019advances}. To enable rapid and flexible prototyping, the FL community lacks a sophisticated system design for scaling an arbitrary number of clients into a fixed small GPU cluster.
    \item \textbf{Partial joining and Stateful clients}. In each round, only a part of the clients that are selected needs to be simulated. This property is usually exploited to relax the hardware requirements by letting executors only load global parameters to execute simulation of selected clients. However, this benefit is hindered when clients are stateful. Specifically, it is unrealistic that each device stores all historical client states and communicates them to other devices to obtain the client state, which requires enormous extra memory and communication overhead.
    \item \textbf{Complexity of migration from simulation to production.} In the current community, the cost of migration is high and requires collaboration from both research team and product team; the simulation is unable to quickly find problems in the deployment; A small error in the code inconsistency will lead to errors in the training algorithm, and eventually a failure of training effective model.
\end{itemize}

\section{Parrot System Design}\label{sec:systemdesign}

\subsection{Overall Architecture}

\begin{figure}[htb!] 
\small
   \setlength{\abovecaptionskip}{0.cm}
   \centering
    \includegraphics[width=1.0\linewidth]{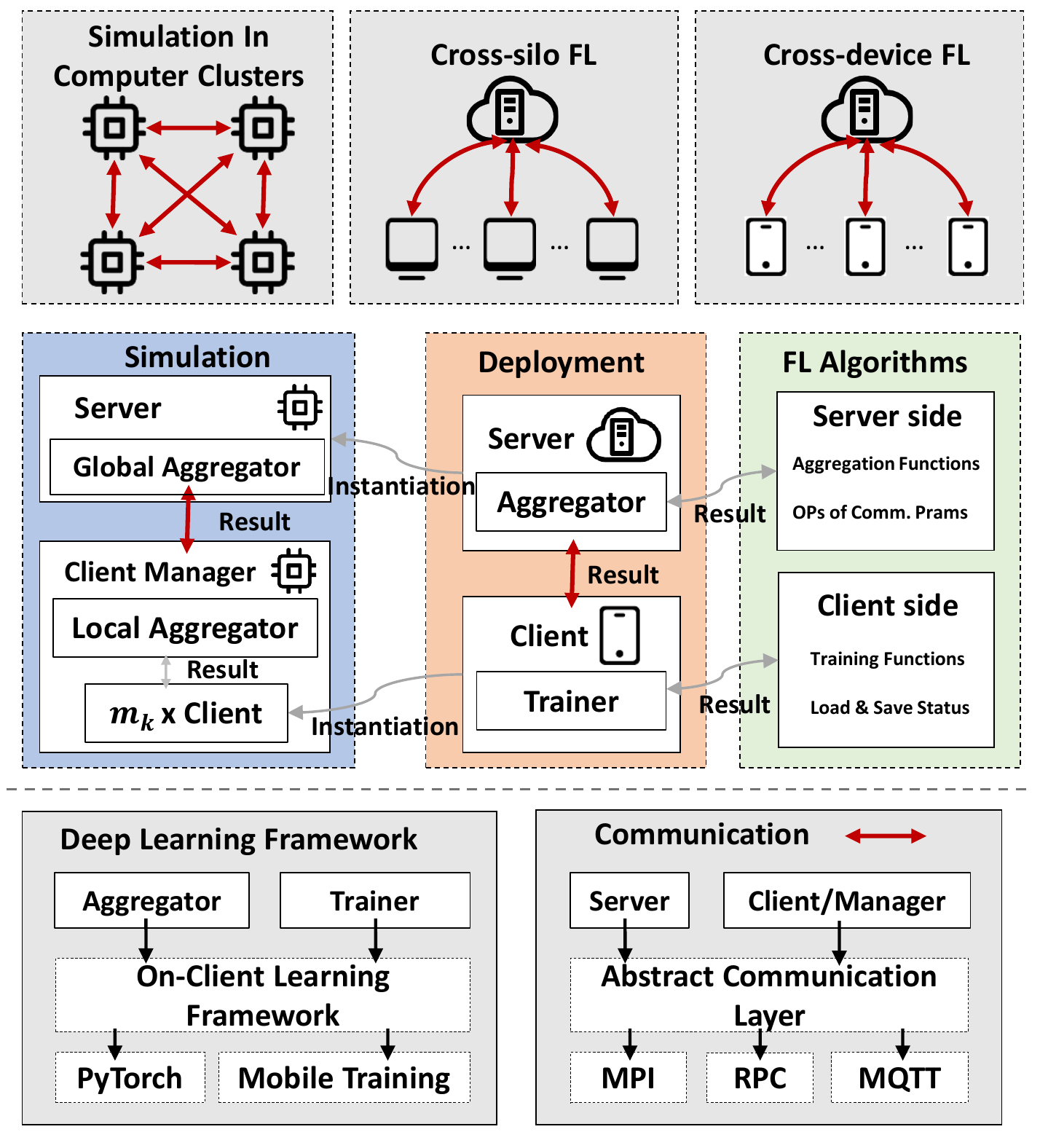}
    \vspace{-0.3cm}
    \caption{Overview of the system design of FedML \texttt{Parrot}.}
    \label{fig:FLframework} 
\vspace{-0.3cm}
% \vspace{-1.0cm}
\end{figure}
\vspace{-0.1cm}

% First, we generalize the FL algorithms into Algorithm~\ref{algo:procedure}. Then, we show our simulation scheme in Algorithm~\ref{algo:simulation}. The overview of our system design is shown in Figure~\ref{fig:FLframework}.

% In this section, we describe the system design of \texttt{Parrot}. 
The FL algorithms can be generalized into Algorithm~\ref{algo:procedure}. Specifically, in almost all FL algorithms, the server needs to communicate some global parameters $\Theta^r$ to clients and collects the client results $\left \{\mathcal{C}_{m, E-1}^{r}|m\in \mathcal{M}^r \right \}$, then conducts some updates. Note that $\Theta^r$ can not only be machine learning model parameters, but can also include some other parameters that are needed by FL algorithms. $\mathcal{C}_{m, j}^{r}$ represents the general returned client result that will be returned to the server. $\mathcal{S}_{m, j}^{r}$ represents the client state required by some FL algorithms. Different from the traditional distributed training schemes, we hope to allow devices to execute multiple tasks in each communication round, thus avoiding the high hardware requirements of deploying each client onto one device.

Algorithm~\ref{algo:simulation} and Figure~\ref{fig:FLframework} show our overall system design. The \textit{Client\_Executes} in Algorithm~\ref{algo:procedure} is seen as a task in Algorithm~\ref{algo:simulation}. In each round, the server manager obtains the selected clients from the original FL server, then conducts scheduling to wisely assign tasks to devices (details in Section~\ref{sec:schedule}). When receiving the assigned tasks, devices execute tasks, locally aggregate client results $\mathcal{C}_{m, j}^{r}$ and save client state $\mathcal{S}_{m, j}^{r}$ one by one. After finishing all tasks, devices return the pre-processed client results $\mathcal{G}_{k}^{r}$ to the server. The server globally aggregates all $\left \{\mathcal{G}_{k}^{r} | k \in \mathcal{K} \right \}$. This local-global aggregation scheme is guaranteed to obtain the same results as the original FL aggregation while reducing communication overheads. This local-global hierarchical aggregation will be specifically introduced in Section~\ref{sec:hierAgg}.

\begin{algorithm}[!t]
	\caption{A General procedure of FL.}
	\label{algo:procedure}
	\small
	\textbf{server input: } initial $\Theta^0$,  maximum communication round $ R  $ \\
	\textbf{client $m$'s input: } local epochs $ E$
	\begin{algorithmic}
		\small
        \STATE \textbf{Server\_Executes:}
            \FOR{each round $ r=0,1, \cdots, R$}
        		\STATE server selects a set of clients $\mathcal{M}^r \subseteq \left \{1, ..., M \right \}$.
                \STATE \small server \textbf{communicates} $\Theta^r$ and to all clients $m \in \mathcal{M}^r$.
        		\FOR{ each client $ m \in \mathcal{M}^{r}$ \textbf{ in parallel do}}
                       \STATE \small $  \mathcal{C}_{m, E-1}^{r+1} \leftarrow $  Client\_Executes $(m, \Theta^{r})$.
        		\ENDFOR
             \STATE $\Theta^{r+1} \leftarrow $ Aggregate($\Theta^r, \left \{\mathcal{C}_{m, E-1}^{r+1}|m\in \mathcal{M}^r \right \}$).
        	\ENDFOR
       \STATE
        \STATE \textbf{Client\_Executes($ m, \Theta^r $):}
            \FOR{each local step $j$ with $j=0,\cdots, E-1 $}
              \STATE Sample raw data $\xi_m \sim \mathcal{D}_m$
              \STATE $\mathcal{C}_{m, j}^{r+1}, \mathcal{S}_{m, j}^{r+1} \leftarrow$  Train $(m, \mathcal{C}_{\xi_m, j}^{r},\mathcal{S}_{m, j}^{r}, \Theta^{r})$
              \ENDFOR
        \STATE \textbf{Return} $\mathcal{C}_{m, E-1}^{r+1}$ to server.
        % \RETURN{ $\theta$ to server}
\end{algorithmic}
\end{algorithm}
\vspace{-0.1cm}

\begin{algorithm}[!t]
	\caption{\texttt{Parrot} Simulation Scheme.}
	\label{algo:simulation}
	\small
	\textbf{server input: } initial $\Theta^0$,  maximum communication round $ R  $ \\
	\textbf{client $m$'s input: } local epochs $ E$
	\begin{algorithmic}
		\small
        \STATE \textbf{Server\_Executes:}
            \FOR{each round $ r=0,1, \cdots, R$}
        		\STATE server selects a set of clients $\mathcal{M}^r \subseteq \left \{1, ..., M \right \}$.
        		\STATE \small  $\left \{\mathcal{M}_{k}^{r} | k \in \mathcal{K} \right \}$ $\leftarrow$ \textbf{Task\_Schedule}($r$, $\mathcal{M}^r$, $\mathcal{T}^r$). 
                \STATE \small server \textbf{communicates} $\Theta^r$ and $\mathcal{M}^r_k$ to each device $k \in \mathcal{K} $.
        		\FOR{ each device $ k \in \mathcal{K}$ \textbf{ in parallel do}}
                       \STATE \small $\mathcal{G}_{k}^{r+1} $ $\leftarrow$  Device\_Executes$(\mathcal{M}^r_k, \Theta^{r})$. 
        		\ENDFOR
             \STATE $\Theta^{r+1} \leftarrow $ GlobalAggregate($\Theta^r, \left \{\mathcal{G}_{k}^{r+1} | k \in \mathcal{K} \right \}$).
        	\ENDFOR
       \STATE
        \STATE \textbf{Device\_Executes}($\small \mathcal{M}^r_k, \Theta^r $):
            \FOR{each $m \in \mathcal{M}_k^r$}
                \STATE $\mathcal{S}_{m,0}^{r} \leftarrow  $ Load\_State$(m)$.
                \STATE $\mathcal{C}_{m, E-1}^{r+1},\mathcal{S}_{m, E-1}^{r+1} \leftarrow $  Client\_Executes$(m, \Theta^{r}, \mathcal{S}_{m, 0}^{r})$.
                \STATE Save\_State$(m, \mathcal{S}_{m, E-1}^{r+1})$.
            \ENDFOR
            \STATE $\mathcal{G}_{k}^{r+1} $ $ \leftarrow $ LocalAggregate($\mathcal{C}_{m, E-1}^{r+1}$)
        \STATE \textbf{Return} $\mathcal{G}_{k}^{r+1} $ to server.
        % \RETURN{ $\theta$ to server}
\end{algorithmic}
\end{algorithm}
\vspace{-0.1cm}

% \subsection{Zero-code Modification: from Simulation to Deployment}
\subsection{Easy Migration: from Simulation to Deployment}
To achieve convenient FL deployment, we design the general API interfaces adapting to the simulation and the real-world distributed deployment. The details of devices executing and the communication are hided from users, enabling them to focus on the design of FL algorithms and models. 

As shown in Figure~\ref{fig:FLframework}, the only extra things that users need to do for implementing FL algorithms in \texttt{Parrot} include (1) specifying the the operations (OPs) on the communicated parameters, e.g. weighted average, summation, simple average, or collected without average. \texttt{Parrot} will conduct aggregation based on the defined OPs; (2) specifying the parameters that needed to be stored if needed (stateful clients). And the server-side and client-side functions can be customized by users like, client selection, local training and server updating etc.. Then, \texttt{Parrot} can automatically schedule client tasks to devices and execute.

\subsection{Sequential Distributed Training}
% The server manager divides $M_p$ tasks into $K$ disjoint sets $\mathcal{M}_1,...,\mathcal{M}_K$, and assign them to $K$ devices respectively. Each computing device $k$ will sequentially conduct the tasks in $\mathcal{M}_k$ one by one like the \texttt{SP} simulation. Devices will instantly conduct one local aggregation after simulating one tasks. When all assigned tasks are completed, each device will return a set of results $\left \{ \mathcal{C}_m | m \in \mathcal{M}_k \right \}$ to the server.

When training each assigned client, devices will first load the client state from the state manager and set the parameters of the current client as the initial states. Then devices conduct the local training according to the user-defined functions. After training, the returned local result $\mathcal{C}_{m, k}^{r}$  will be locally aggregated according to the user-defined OPs. Then the new results $\mathcal{G}_{ k}^{r}$ of device $k$ will be returned to the server for further aggregation.

\subsection{State Management for Stateful FL Clients}

When algorithms have stateful clients, simulating large-scale FL needs to store client state of massive sizes. Assuming the size of client state is $s_d$, the total client state occupies $s_dM$ memory. It is impossible to load it into the CPU or GPU memory. Thus, we build a state manager to automatically load and save client state in and from disks. When devices begin to simulate client $k$, it can load the according client state from the disk. After finishing training, this client updates the state and the manager automatically saves it into the disk.

\section{Parrot System Optimization}\label{sec:perfOpt}

In this section, we introduce the system performance optimization of \texttt{Parrot}. Our system optimization aims to reduce communication overheads, and schedule clients to achieve the minimization of the total training time.

\subsection{Necessity of Scheduling}
In the FL, clients have very high heterogeneous dataset sizes so the training time of different clients is quite different. This causes the well-known straggler problem, where the server must wait for the slowest executor to finish its tasks, causing long training time and very low GPU utilization. Besides, the training devices may not be homogeneous. Users may have machines equipped with different CPUs or GPUs. Or, devices may be used to conduct other tasks, causing unstable computing performance. These factors can all cause the straggler problem.

\subsection{Hierarchical Aggregation}\label{sec:hierAgg}

Many FL algorithms average parameters, gradients, or the differences of machine learning models from $M_p$ clients~\citep{mcmahan2017communication,fedprox,wang2020tackling,reddi2020adaptive,acar2021federated,chen2021bridging}. Assuming the parameters that need to be averaged (AVG. Params.) is of size $s_a$, the total communicating size in one round is $s_aM_p$. And the server needs to sum AVG. Params. for $M_p-1$ times.

We note that this average process can be decomposed into the local and global average, reducing the communication size and the aggregation cost of the server. Then, we modify the aggregation operation at the FL server into a local-global  hierarchical aggregation scheme. Specifically, devices will locally average $|\mathcal{M}^r_k|$ AVG. Params., and server globally aggregate the $K$ results from devices. Thus, the communication cost is reduced from $s_aM_p$ to $s_aK$, communication trips are reduces from $M_p$ to $K$, and server only need to conduction sum operations for $K-1$ times.

However, some algorithms~\citep{fedmask,exploitsharedRep,FedDF} need to communicate parameters that should be collected at the server but not averaged (Special Params.). For these parameters of size $s_e$, devices wrap them into a message and send to the server. The server collects them in the same way as the FL server. In this case, we can only reduce the communication trips but not the total communication size $s_eM_p$. Luckily, there are not many FL algorithms with Special Params. of large size. Because the server needs extra $s_eM_p$ or $s_eM$ memory to store and use them. When $M_p$ and $M$ are large, this gives the server too much system overhead.

\subsection{Quantifying the Per-client Running Time}\label{sec:estimation}

The straggler problem, i.e. waiting for the slowest device, severely slows down the system efficiency. Thus, we hope to balance the workload (training time) between devices, trying to make devices finish their tasks at almost the same time. To implement this, firstly, we should accurately estimate the workload of tasks on devices.

% ~\citep{CommModel,MGWFBP}
\textbf{Workload Model.}
Because we hope to balance the running time of devices, we define the workload of a task $\pi_m$ on device $k$ as $T_{m,k}$. Local training on client $k$ mainly includes parts: (1) loading dataset; (2) forward propagation; (3) backward propagation; (4) updating model. The total running time is dominated by (2) and (3), which usually increase almost linearly with the dataset size. Thus, we build a simple linear model to estimate the running time, which is widely used in distributed training:
\begin{equation}\label{eq:TimeModel}
T_{m} = N_m * t^{sample} + b,
\end{equation}
in which $N_m$ represents the size of dataset $\mathcal{D}_m$ on client $m$, $t^{sample}$ the processing time of one data sample, and $b$ the constant time of conducting one task by device $k$. 

The $t^{sample}$ includes all processing time of (1)$\sim$(4) per sample. And the $b$ is constant, as there are some procedures that have a time cost independent of the dataset, including moving the model between CPU memory and GPU memory, setting client model parameters as the server model parameters, etc.

\textbf{Heterogeneous Hardware Resources.}
Considering the heterogeneity of computing devices, the processing time on different devices may be highly variable. To more accurately model the workload, we modify the Equation~\ref{eq:TimeModel} as:
\begin{equation}\label{eq:TimeModelDifDevice}
T_{m,k} = N_m * t_{k}^{sample} + b_{k}.
\end{equation}
Now, the workload model considers the heterogeneity of devices, to better schedule the tasks to clients.

\textbf{Estimation.}
Predicting the running time based on the hardware resources, dataset, model architecture, and hyper-parameters like batch size, and number of iterations is difficult and costs a lot of time to achieve accurate estimation~\citep{}. Thus, we prefer to use the historical running time to fit our workload model.

During the training process, each device $k$ records the running time of conducting tasks $\left \{ \hat{T}_{m,k}^r|m \in \mathcal{M}_k^r \right \}$ at round $r$, and sends them back to the server together with the results $\left \{ \mathcal{C}_{m,E-1}^r|m \in \mathcal{M}_k^r \right \}$.

When conducting scheduling at round $r$, the server will use all $\mathcal{T}^r = \left \{ \hat{T}_{m,k}^i|m \in \mathcal{M}_k^r, k \in \mathcal{K}, i \in [0,...,r-1 ] \right \}$ to fit the workload model, i.e. Equation~\ref{eq:TimeModelDifDevice}, obtaining the estimated parameters $t_{k}^{sample}$ and $b_k$.

\subsection{Efficient Task Scheduling for Load Balancing}\label{sec:schedule}

% $\mathcal{M}^r_1,...,\mathcal{M}^r_K$ $\leftarrow$ \textbf{Task\_Schedule}($\mathcal{M}^r$, $\mathcal{T}^r$).

% \STATE \textbf{Task\_Schedule}($r$, $\mathcal{M}^r$,  $\mathcal{T}^r$):
\begin{algorithm}[!t]
	\caption{Task Scheduling.}
	\label{algo:schedule}
	\small
	\textbf{input: } current round $r$, selected clients $\mathcal{M}^r$, history running time $\mathcal{T}^r$
	\begin{algorithmic}
        \IF {$r \le R_w  $}
            \STATE Uniformly divides $\mathcal{M}^r$ into $\left \{\mathcal{M}_{k}^{r} | k \in \mathcal{K} \right \}$ with similar $|\mathcal{M}^r_k|$
        \ELSE
            \STATE $t_{k}^{sample}, b_k \leftarrow$ \textbf{Estimate\_Workload($\mathcal{T}^r$)}
            \STATE initialize task sets $\left \{\mathcal{M}_{k}^{r}= \emptyset| k \in \mathcal{K} \right \}$.
            \STATE initialize  $\left \{w_k=0 | k \in \mathcal{K} \right \}$.
            \STATE $\mathcal{M}^{s}_{sort} \leftarrow$ \textbf{Sort}$\left\{ N_m |m \in \mathcal{M}^r \right \} $
    		\FOR{ each client $ m \in \mathcal{M}^{r}_{sort}$ }
                % \STATE \small $T_{m,k} = N_m * t_{k}^{sample} + b_{k}$.
                \STATE find $k^\star$ by Equation~\ref{eq:kstar}.
                \STATE put $m$ into $ \mathcal{M}^r_{k^\star}$.
                \STATE $w_{k^\star} = w_{k^\star} + N_m * t_{k^\star}^{sample} + b_{k^\star}$.
    		\ENDFOR
    	\ENDIF
        \STATE \textbf{Return} $\mathcal{M}^r_1,...,\mathcal{M}^r_K$.
      \STATE
        \STATE \textbf{Estimate\_Workload($\mathcal{T}^r$):}
        \IF {Using Recent running time }
            \STATE $ t_{k}^{sample}, b_k \leftarrow$ Linear Regression on Equation~\ref{eq:TimeModelDifDevice} with ($\mathcal{T}_\tau^r$).
        \ELSE
            \STATE $ t_{k}^{sample}, b_k \leftarrow$ Linear Regression on Equation~\ref{eq:TimeModelDifDevice} with ($\mathcal{T}^r$).
        \ENDIF
        \STATE \textbf{Return} $t_{k}^{sample}, b_k$.
\end{algorithmic}
\end{algorithm}
\vspace{-0.3mm}

The optimization goal of task scheduling is to minimize the total training time. Because devices conduct simulation in parallel, the total training time is decided by the device that needs the longest training time. Here we represent the training time of device $k$ as the workloads $w_k$. Ignoring the extra switching time between tasks, with the workload model of Equation~\ref{eq:TimeModelDifDevice}, we have accumulated workloads $w_k = \sum_{m \in \mathcal{M}_k } T_{m,k}$. Then, the estimated training time of one round is $ \text{max}(w_k| k \in \mathcal{K})$. Thus, the optimization problem can be formulated as:
\begin{equation}
    \mathop{\min}_{\mathcal{M}_1,...,\mathcal{M}_K} \ \text{max}(\left\{ \sum_{m \in \mathcal{M}_k } T_{m,k} |k \in \mathcal{K} \right \}) .
\end{equation}

We design an efficient task scheduling method in Algorithm~\ref{algo:schedule}, with computing complexity $O(KM_p)$. We firstly sort the $\left\{ N_m |m \in \mathcal{M}^r \right \}$ in the descending order as $\mathcal{M}^{r}_{sort}$. Then we traverse the $m \in \mathcal{M}^{r}_{sort}$ to gradually find the solution $\mathcal{M}_1,...,\mathcal{M}_K$. When scheduling a new task $m$, we assign it to device $k^\star$ to fulfill:
\begin{equation}\label{eq:kstar}
    k^\star = \mathop{\arg\min}_{k \in \mathcal{K}} \ \text{max}(\left\{w_k + N_m * t_{k}^{sample} + b_{k} |k \in \mathcal{K}\right \}).
\end{equation}
Task of simulating client $m$ will be put into $ \mathcal{M}^r_{k^\star}$. After traversing the $\mathcal{M}^{r}_{sort}$, the server can get the $\mathcal{M}_1,...,\mathcal{M}_K$ and send them to devices.

% By viewing this scheduling problem as a dynamic programming problem, we firstly sort the $\left\{ N_m |m \in \mathcal{M}^r_s \right \}$ in the descending order as $\mathcal{M}^{r}_{sort}$.

\textbf{Tackling Dynamic Hardware Environments.}
Usually, researchers and engineers use computing resources with other users simultaneously. Such an unstable environment makes $t_{k}^{sample}$ and $b_k$ vary during the training process. Thus, the estimation based on all historical running time becomes inaccurate. 

To this end, assuming the hardware environment is relatively stable in a short recent period, we propose to discard the early recorded information and only exploit the running time in a recent Time-Window for estimation. Users can define a time threshold $\tau$ based on the instability of their environments. Then  $\mathcal{T}^r_\tau = \left \{ \hat{T}_{m,k}^i|m \in \mathcal{M}_k^r, k \in \mathcal{K}, i \in [r-\tau,...,r-1 ] \right \}$ will be used for estimation. We call this with the scheduling as Time-Window scheduling.

\subsection{Complexity Analysis}\label{sec:complexity}
% In this section, we will analyse the requirements of hardware of different simulation schemes. Here, we use $M$ to represent the number of total clients, $M_p$ the number of selected clients per round, $K$ the number of devices (executors), $s_m$ the size of memory that is needed to simulate one client. $s_a$ represents the size AVG. parameters. $s_e$ is the size of special parameters. $s_d$ is the size of client state, which is only needed by algorithms with stateful clients. The complexity of different simulation schemes is summarized in Table~\ref{tab:complexity}.
In this section, we will analyse the requirements of hardware of different simulation schemes. The definitions of notations and the complexity of different simulation schemes is summarized in Table~\ref{tab:complexity}.

\textbf{CPU/GPU Memory \& Disk Cost.}
The \texttt{SP} simulation only uses one device to simulate FL. Storing all models and client state on it needs memory of $O(s_mM +s_dM)$ size. Loading client model and state from disk can help it reduce the memory cost as $O(s_m + s_d)$, but with extra dist cost $O(s_mM_p + s_dM)$.

As shown in Figure~\ref{fig:FLsimulation}, the \texttt{RW Dist.} needs to deploy all clients, thus there should be total $O(s_mM + s_dM)$ size of memory to store and train client models. With the state managing, un-selected clients need not to load state from the disk, thus the memory is reduced by $O(s_mM + s_d(M-M_p))$, and the disk cost increases to $O(s_dM)$ accordingly. 

In the \texttt{SD Dist.}, devices only need to load the global model and conduct simulation of selected clients, instead of deploying all clients. This part needs size of $O(s_mM_p)$. Because the client state of un-selected clients cannot be discarded, devices need $O(s_mM)$ size of memory to store them. Similar with \texttt{RW Dist.}, loading client state in this disk can reduce memory cost as $O(s_mM_p + s_dM_p)$.

The \texttt{SD Dist.} and \texttt{Parrot} both assign tasks to $K$ devices to simulate clients one by one. Thus they only need memory size of $O(s_mK)$ to train client models and $O(s_dM)$ to store client state. With state manager, $K$ devices can load client state when needed. Thus the total memory cost is reduced as $O(s_mK + s_dK)$.

Note that for all schemes, the disk cost for client states cannot be further reduced. Thus, the size of disk may be the bottleneck of simulating large-scale FL with stateful clients.

\textbf{Communication cost.}
As shown in Figure~\ref{fig:FLsimulation} and ~\ref{fig:FLtimelines}, the server in \texttt{RW Dist.} and \texttt{SD Dist.} need to sends parameters to $M_p$ clients, and receive $M_p$ results from them. Thus the communication trips are $O(M_p)$, and communication size is $O(s_aM_p + s_eM_p)$.

For the \texttt{FA Dist.}, the server assigns only one task and parameters to clients when they are available. Thus, the \texttt{FA Dist.} actually has the same communication trips and communication size with \texttt{SD Dist.}.

Benefitted from the hierarchical aggregation of \texttt{Parrot}, devices only need to send the results once rather than multiple times. Thus, there are only $O(K)$ communication trips and $O(s_aK + s_eM_p)$ communication size. Here, because those special parameters are designed to be collected but not averaged on the server, the communication size $O(s_eM_p)$ cannot be optimized more.

\textbf{Time Costs of Workload Estimation.}
If one uses all the history data of running time to fit the workload model~\ref{eq:TimeModelDifDevice}. For each workload model of device $k$, the number of data points is $rM_p/K$ at communication round $r$. Because there are only $2$ features of each workload model, i.e. $t_k^{sample}$ and $b_k$, the complexity of linear regression on each model is $O(4(rM_p)^2/K^2)$. If one needs to conduct large-scale FL for a long time, he or she can cut off former history data like the Time-Window scheduling in Section~\ref{sec:estimation} for reducing the computation complexity.

\textbf{Time Costs of Scheduling.}
In our scheduling algorithm, there is a traversal process of $\mathcal{M}^r$. For each $ m \in \mathcal{M}^r$, we need to find a $k^\star$ to minmax the accumulated workloads, which needs $O(K)$ times of searching. Thus, the computational complexity is $O(KM_p)$. We will show in the experiments that the time of workload estimation and scheduling is much less that the training time.

\begin{figure*}[ht!]
    \subfigbottomskip=-1pt
    \subfigcapskip=1pt
  \centering
% \!\!\!\!\!\!\!\!
     \subfigure[Algorithms Without special Params. and stateful clients. ]{\includegraphics[width=0.24\linewidth]{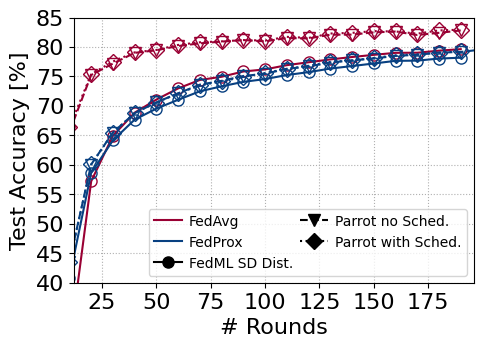}\label{fig:TestAccFMNISTclass_algo}}
     \subfigure[Algorithms With special Params. but without stateful clients. ]{\includegraphics[width=0.24\linewidth]{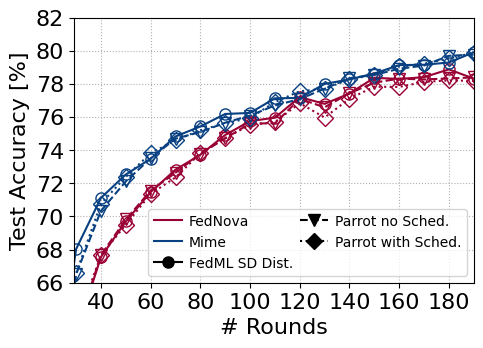}\label{fig:TestAccFMNISTspe_param_algo}}
     \subfigure[Algorithms With both special Params. stateful clients. ]{\includegraphics[width=0.24\linewidth]{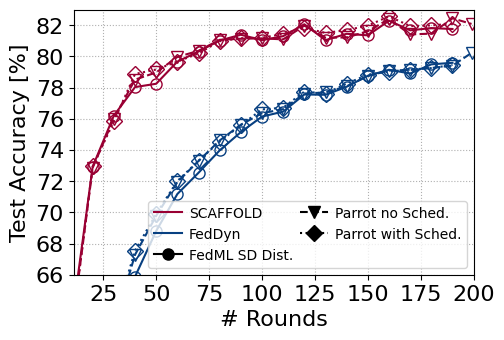}\label{fig:TestAccFMNISTstateful_algo}}
     \subfigure[Running time per round. ]{\includegraphics[width=0.24\linewidth]{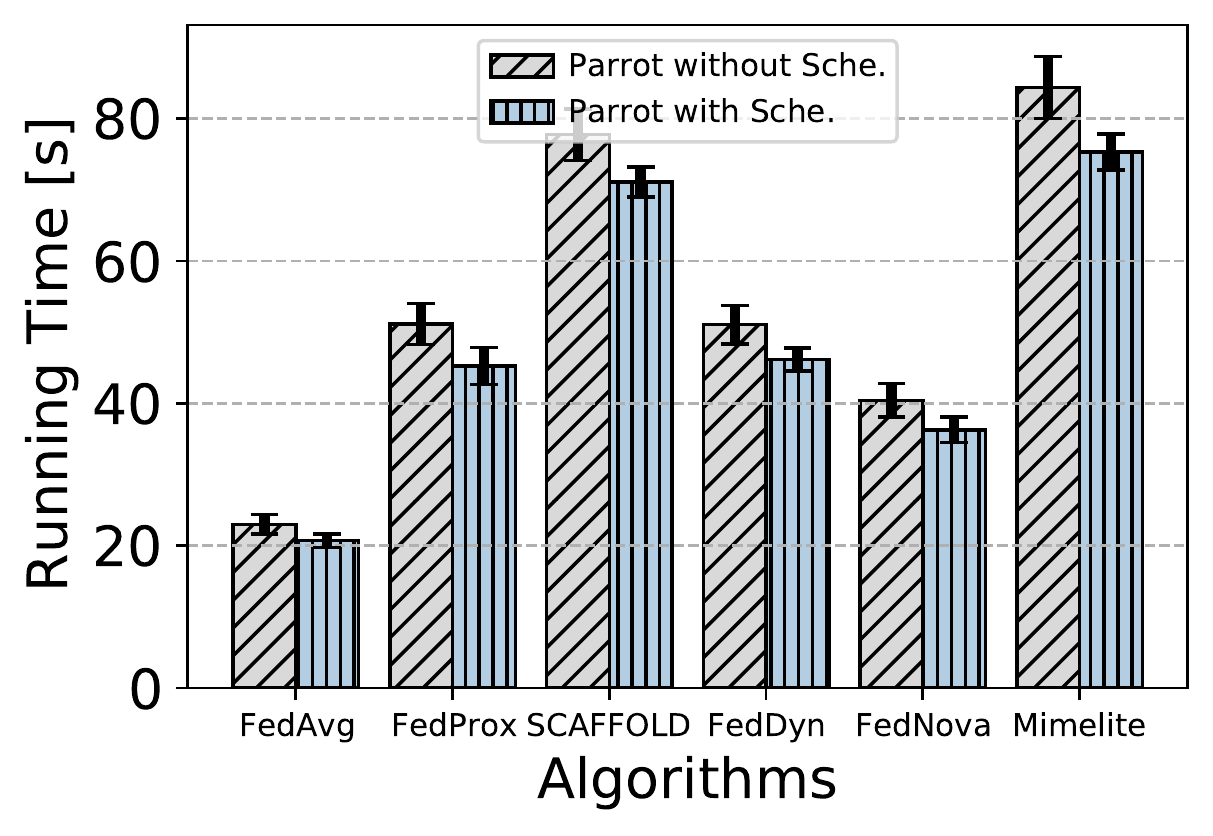}\label{fig:RunTimeFMNISTAlgo}}
    \caption{Test accuracy and running time per round with different FL algorithms.}
    \label{fig:TestAccFMNISTAlgo}
\vspace{-0.3cm}
% \vspace{-0.5cm}
\end{figure*}  

% \begin{figure}[htb!] 
% \small
%   \setlength{\abovecaptionskip}{0.cm}
%   \centering
%     \includegraphics[width=0.6\linewidth]{runtime_bar/RunTimeRound_e_fem_scal_algo.pdf} 
%     \caption{Running time per round of FEMNIST.} 
%     \label{fig:RunTimeFMNISTAlgo} 
% \end{figure}
% \vspace{-0.1cm}

\begin{figure}[ht!]
    \subfigbottomskip=-1pt
    \subfigcapskip=1pt
  \centering
     \subfigure[FEMNIST on Cluster A. ]{\includegraphics[width=0.46\linewidth]{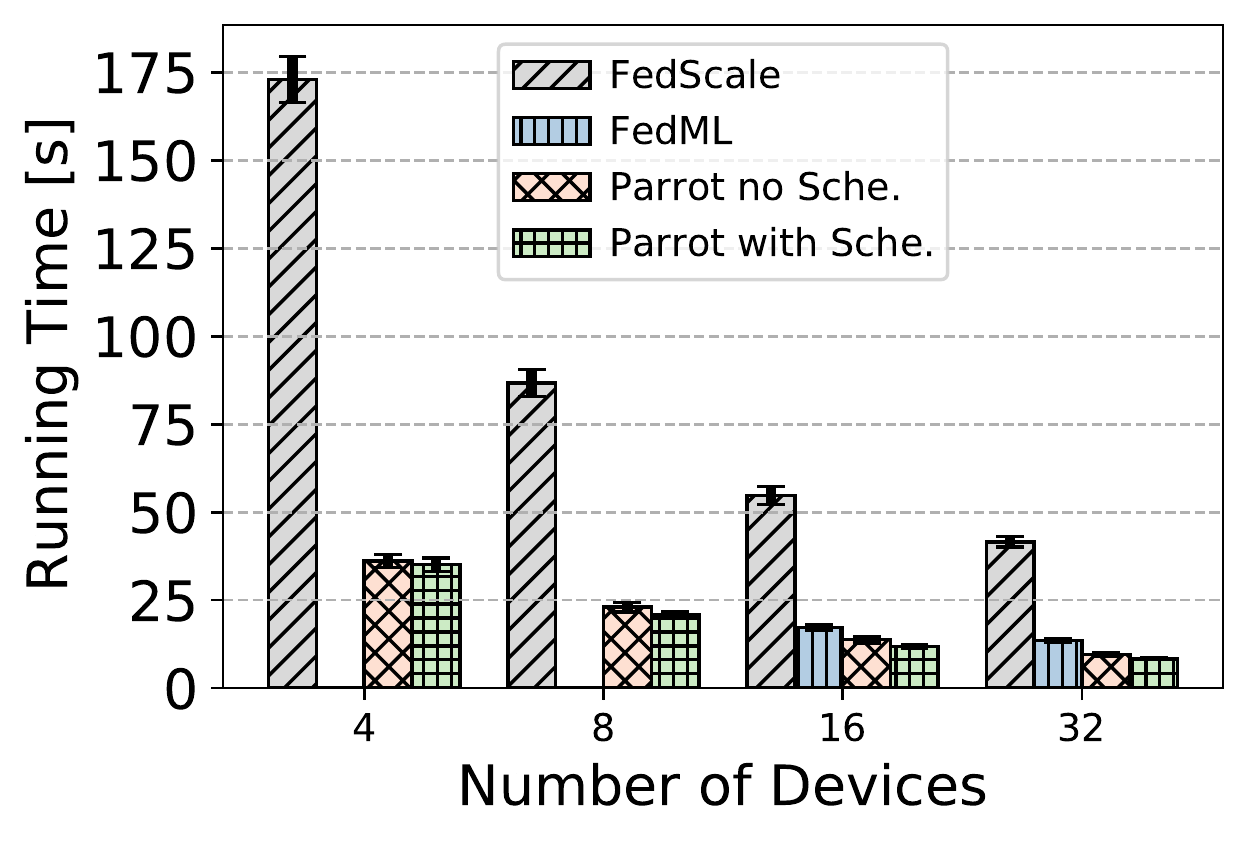}}
     \subfigure[ImageNet(a) on Cluster A. ]{\includegraphics[width=0.49\linewidth]{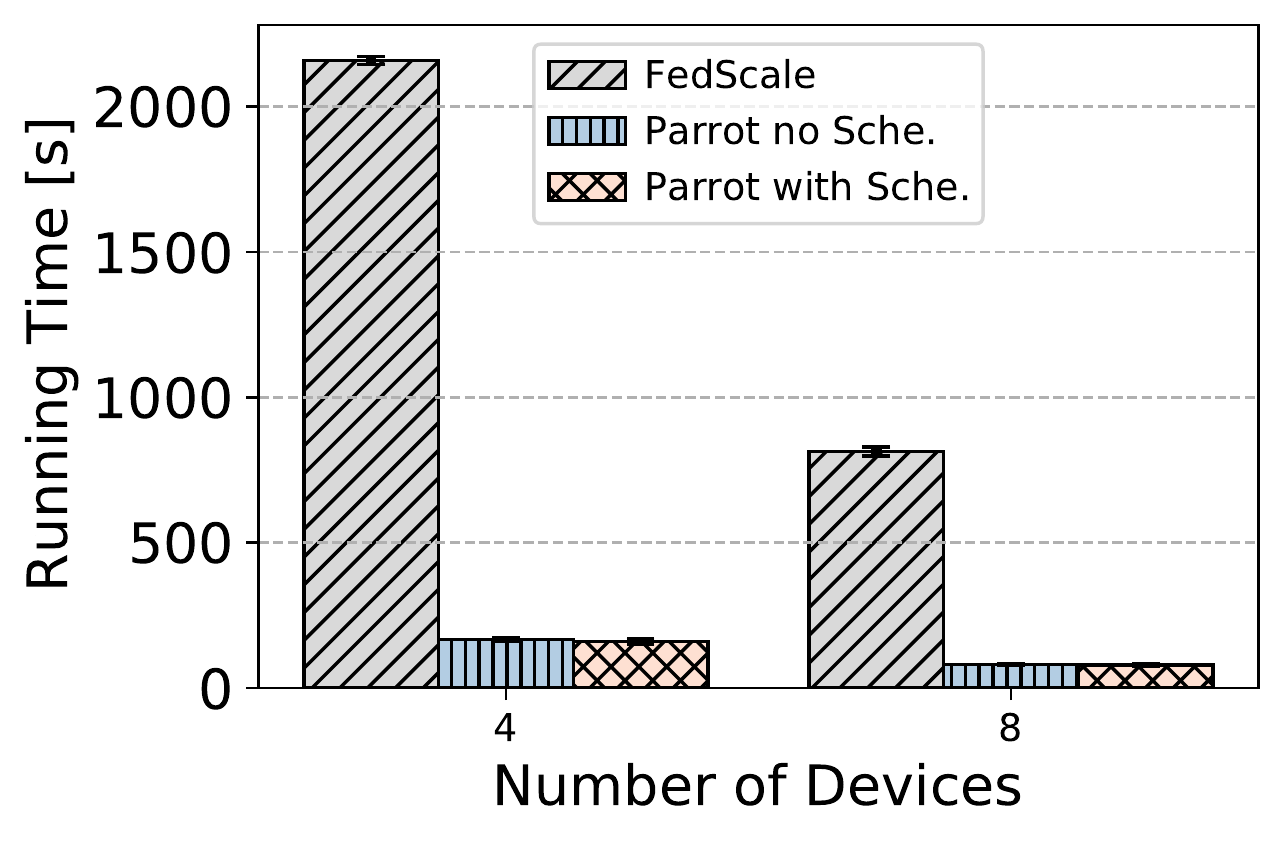}}
     \subfigure[FEMNIST on Cluster B.  ]{\includegraphics[width=0.32\linewidth]{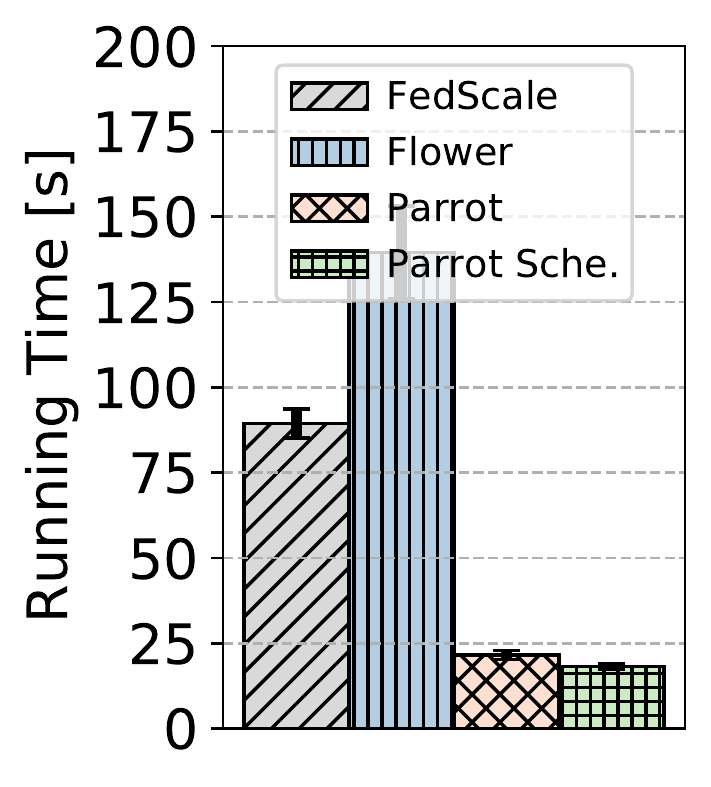}}
     \subfigure[Reddit on Cluster B. ]{\includegraphics[width=0.30\linewidth]{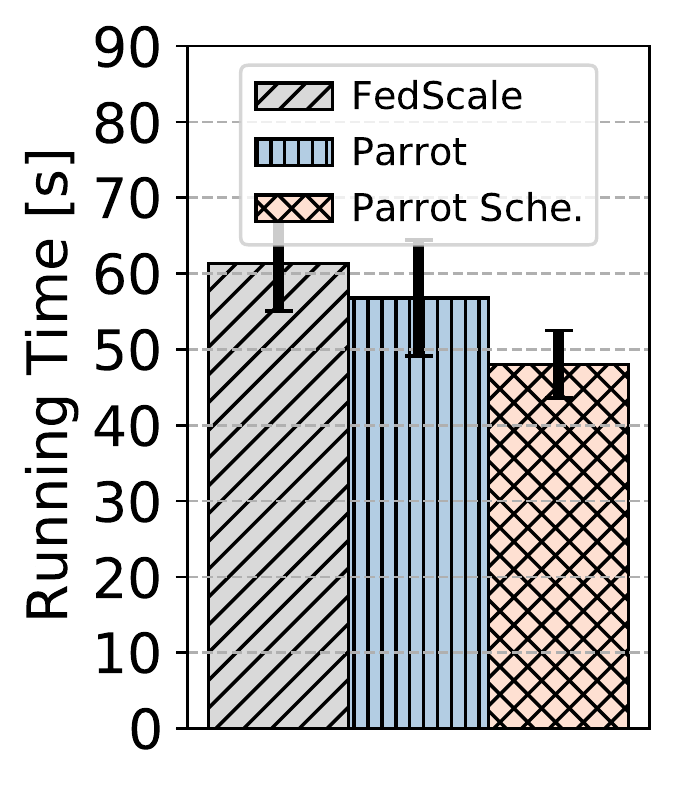}}
     \subfigure[ImageNet(b) on Cluster A. ]{\includegraphics[width=0.33\linewidth]{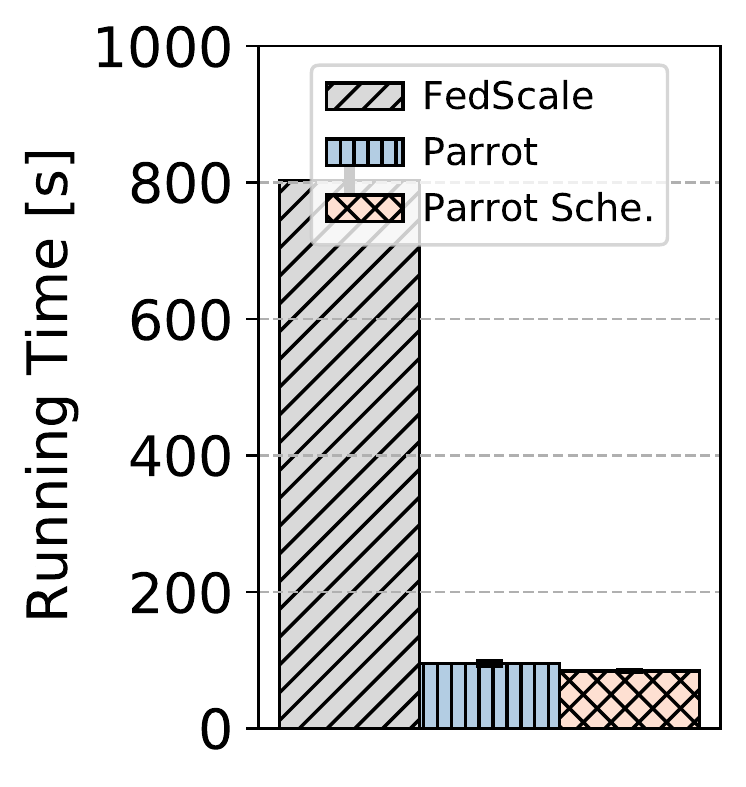}}
     \caption{Running time per round of different FL frameworks with different number of devices.}
    \label{fig:EstimationError_C100_Framework}
\vspace{-0.3cm}
% \vspace{-0.5cm}
\end{figure}
\vspace{-0.3cm}

\begin{figure*}[ht!]
    \subfigbottomskip=-1pt
    \subfigcapskip=1pt
  \centering
% \!\!\!\!\!\!\!\!
     \subfigure[ ]{\includegraphics[width=0.24\linewidth]{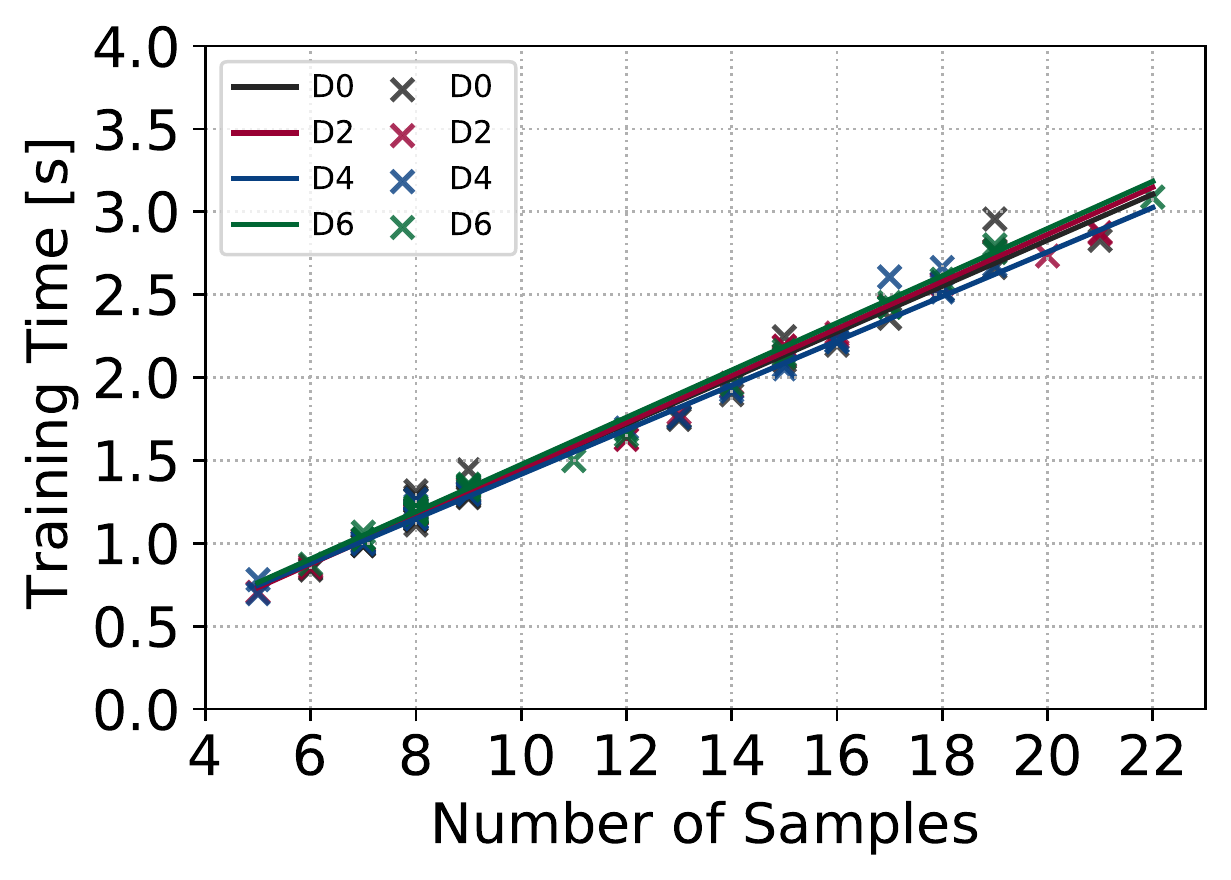}}
     \subfigure[ ]{\includegraphics[width=0.23\linewidth]{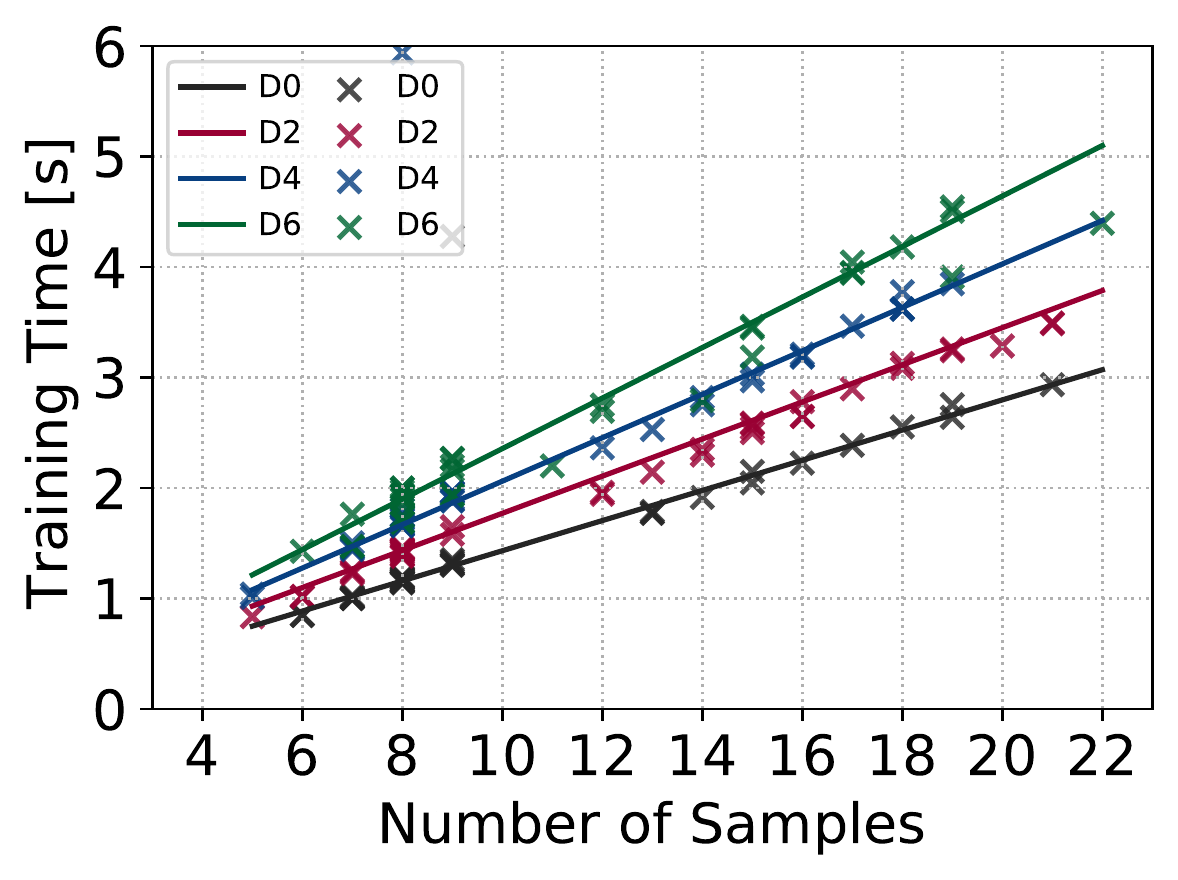}}
     \subfigure[ ]{\includegraphics[width=0.245\linewidth]{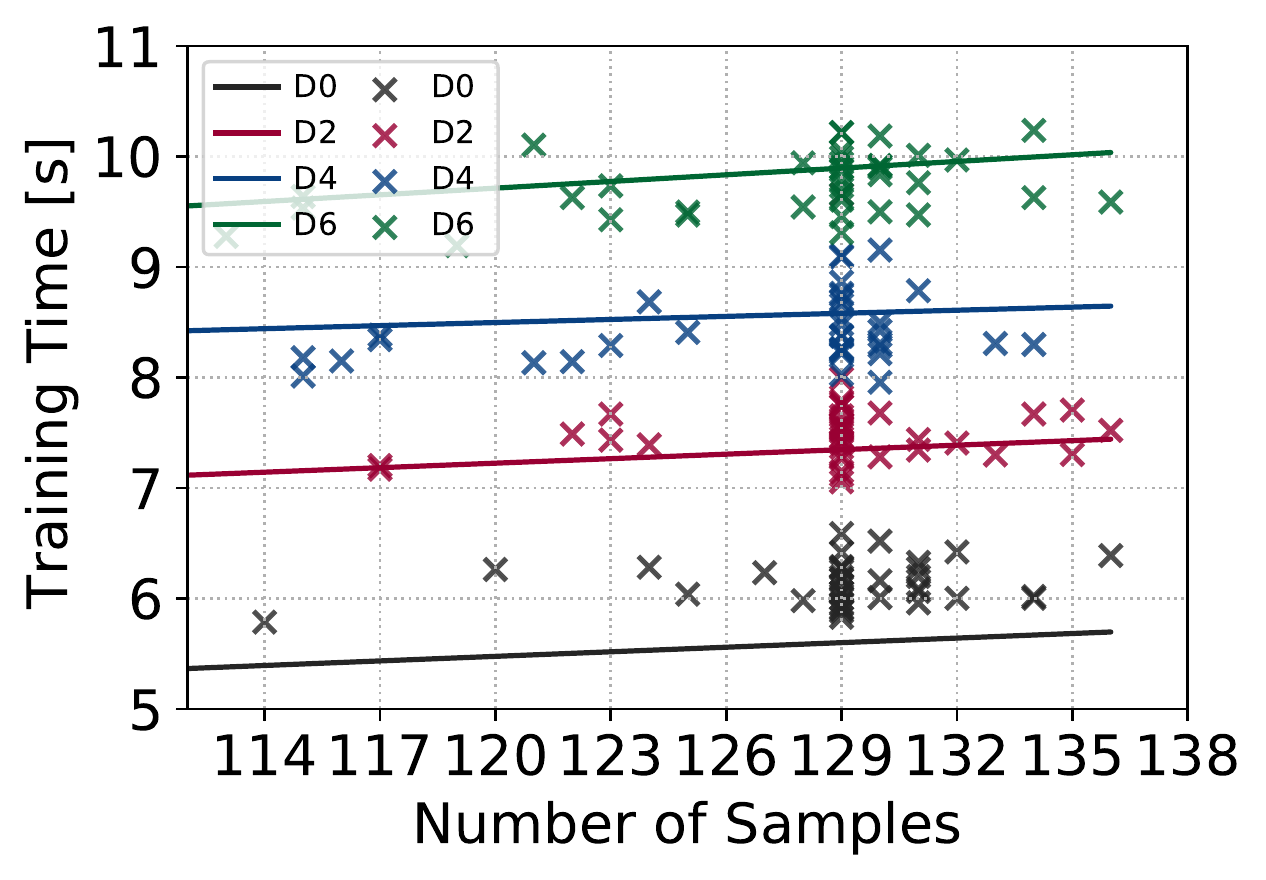}}
     \subfigure[]{\includegraphics[width=0.23\linewidth]{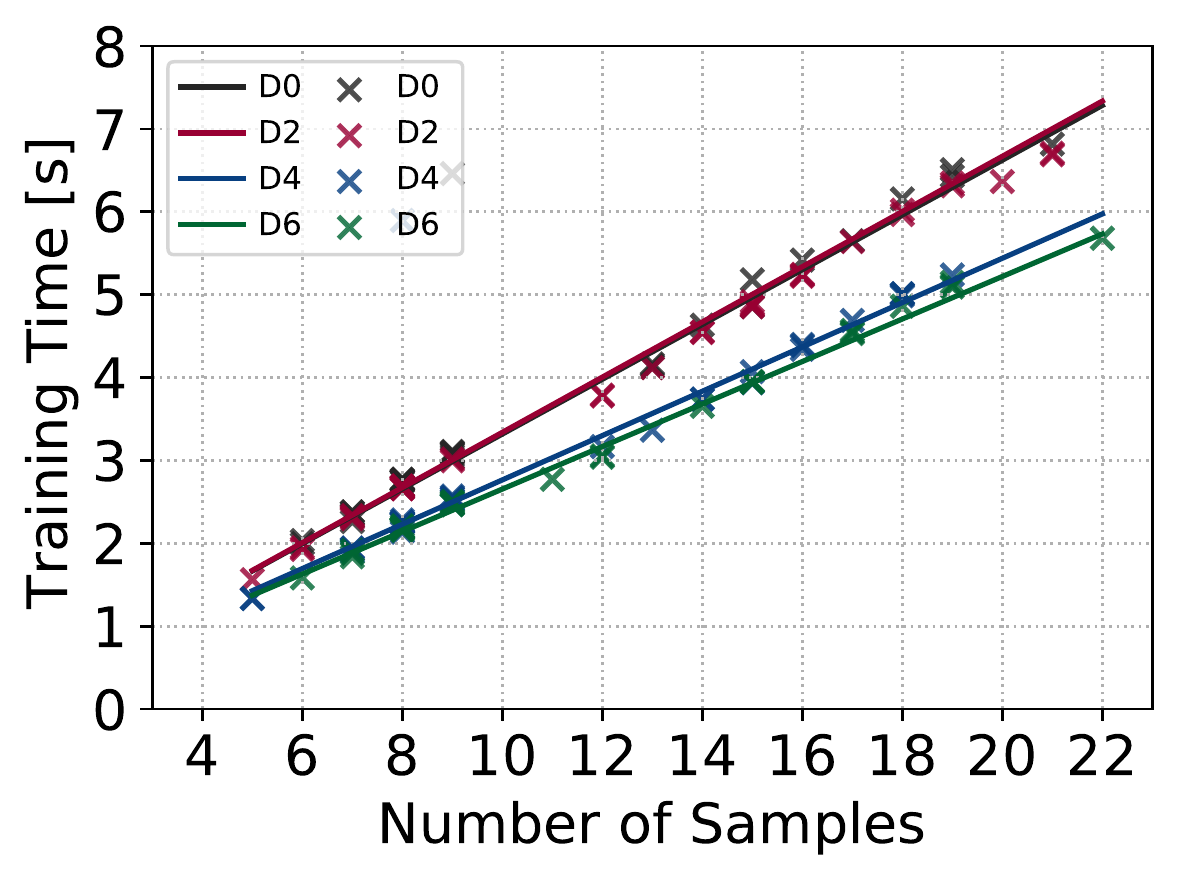}}
     \caption{Running times of training FEMNIST with 8 devices with different environments. We show the estimated workload model and some sampled running times. Here we only choose 4 devices to show for more clear visualization. (a) On Homogeneous GPU Cluster A. (b) FEMNIST, $K=8$, Homogeneous GPU Cluster A, simulating heterogeneous GPUs. (c) ImageNet, $K=8$, Homogeneous GPU Cluster A, simulating heterogeneous GPUs. (d) FEMNIST, $K=8$, Heterogeneous GPU Cluster C.}
    \label{fig:EstimationHomoEnv}
\vspace{-0.3cm}
% \vspace{-0.5cm}
\end{figure*}

% \begin{table*}[ht!]
% \centering
% \caption{GPU Memory Costs (MB).} 
% \vspace{1pt}
% \footnotesize{
% \begin{tabular}{lcccc}
% \hline
% \multicolumn{1}{c}{} & \multicolumn{2}{c}{FEMNIST}      & \multicolumn{2}{c}{ImageNet}       \\ \hline
%                      & $M_p=100, K=8$ & $M_p=100, K=16$ & $M_p=1000, K=8$ & $M_p=1000, K=16$ \\ \hline
% SP                   & 1,122          & 1,112           & 3,305           & 3,305            \\
% SD Dist.             & 112,200        & 112,200         & 3,305,000       & 3,305,000        \\
% FA Dist. Parrot       & 8976           & 17,952          & 26,440          & 52,880           \\ \hline
% \end{tabular}
% }
% \label{tab:GPUmemoryCost}
% \end{table*}

% \begin{figure}[ht!]
%     % \setlength{\abovedisplayskip}{-2pt}
%     % \setlength{\abovecaptionskip}{-2pt}
%     \subfigbottomskip=-1pt
%     \subfigcapskip=1pt
%   \centering
% % \!\!\!\!\!\!\!\!
%      \subfigure[FEMNIST with $M_p=100$. ]{\includegraphics[width=0.48\linewidth]{estimation/EstimateError_e_fem_scal.pdf}}
%      \subfigure[ImageNet with $M_p=100$. ]{\includegraphics[width=0.48\linewidth]{estimation/EstimateError_e_img_scal.pdf}}
%     \caption{The error of workload estimation.}
%     \label{fig:EstimationError_C100_SCALE}
% \vspace{-0.3cm}
% % \vspace{-0.5cm}
% \end{figure}

\begin{figure}[ht!]
    \subfigbottomskip=-1pt
    \subfigcapskip=1pt
  \centering
% \!\!\!\!\!\!\!\!
     \subfigure[FEMNIST with $M_p=100$. ]{\includegraphics[width=0.48\linewidth]{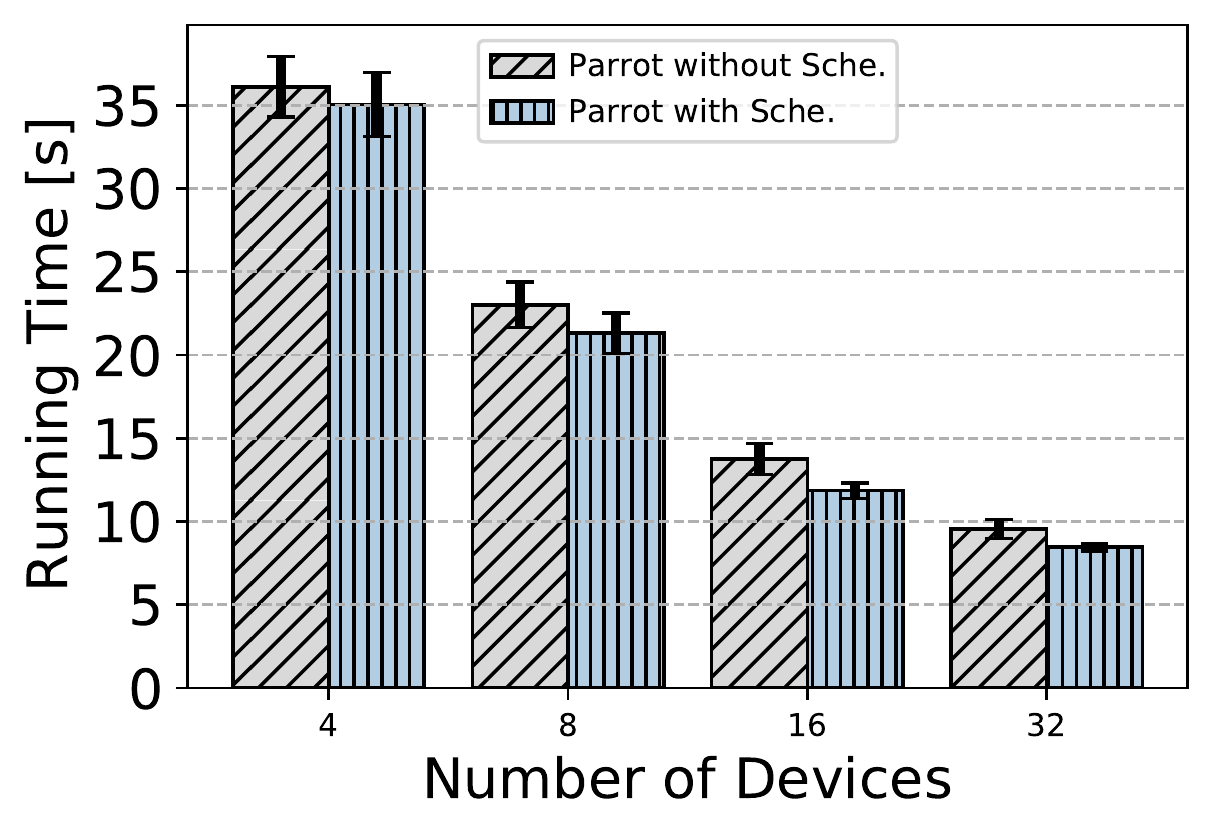}}
     \subfigure[ImageNet with $M_p=100$. ]{\includegraphics[width=0.48\linewidth]{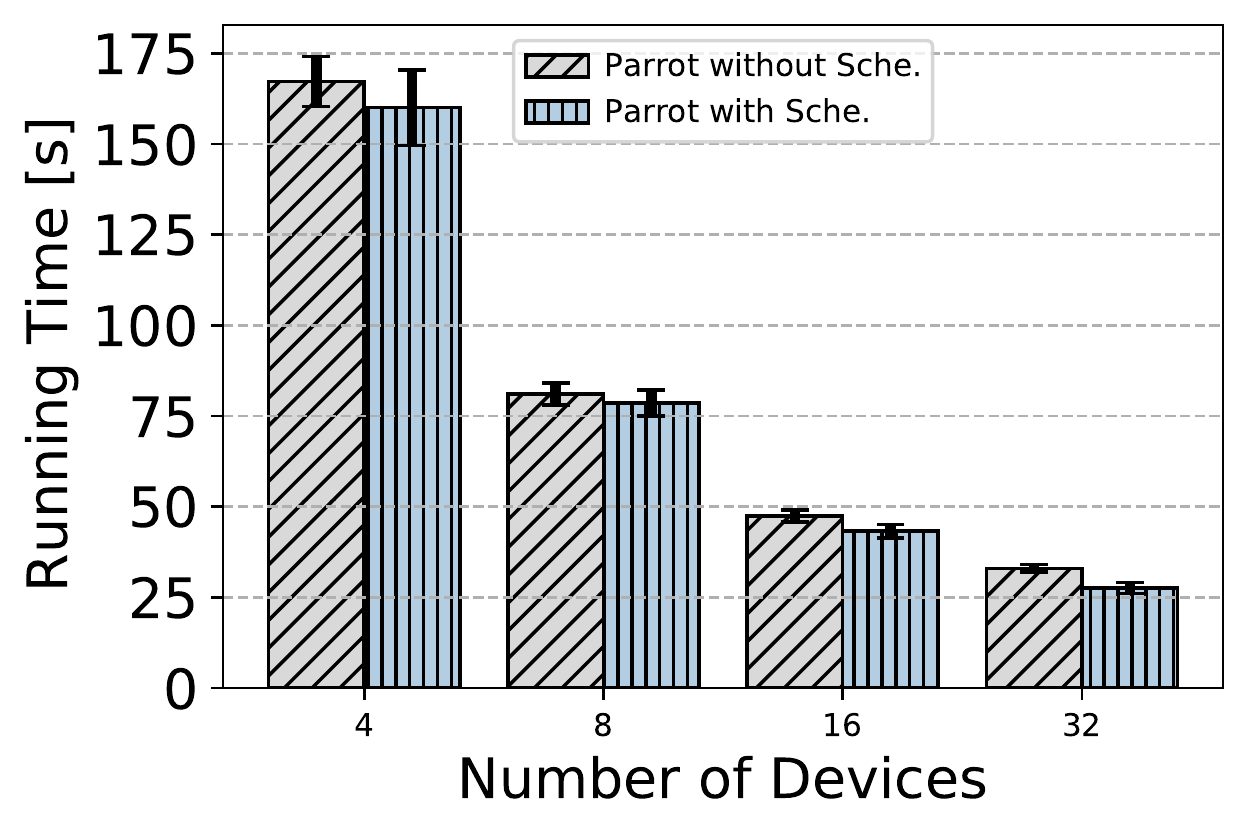}}
    \caption{Running time per round with different number of devices.}
    \label{fig:RunTimeRound_C100_SCALE}
\vspace{-0.3cm}
% \vspace{-0.5cm}
\end{figure}

\begin{figure}[ht!]
    \subfigbottomskip=-1pt
    \subfigcapskip=1pt
  \centering
% \!\!\!\!\!\!\!\!
     \subfigure[FEMNIST with $M_p=100$. ]{\includegraphics[width=0.48\linewidth]{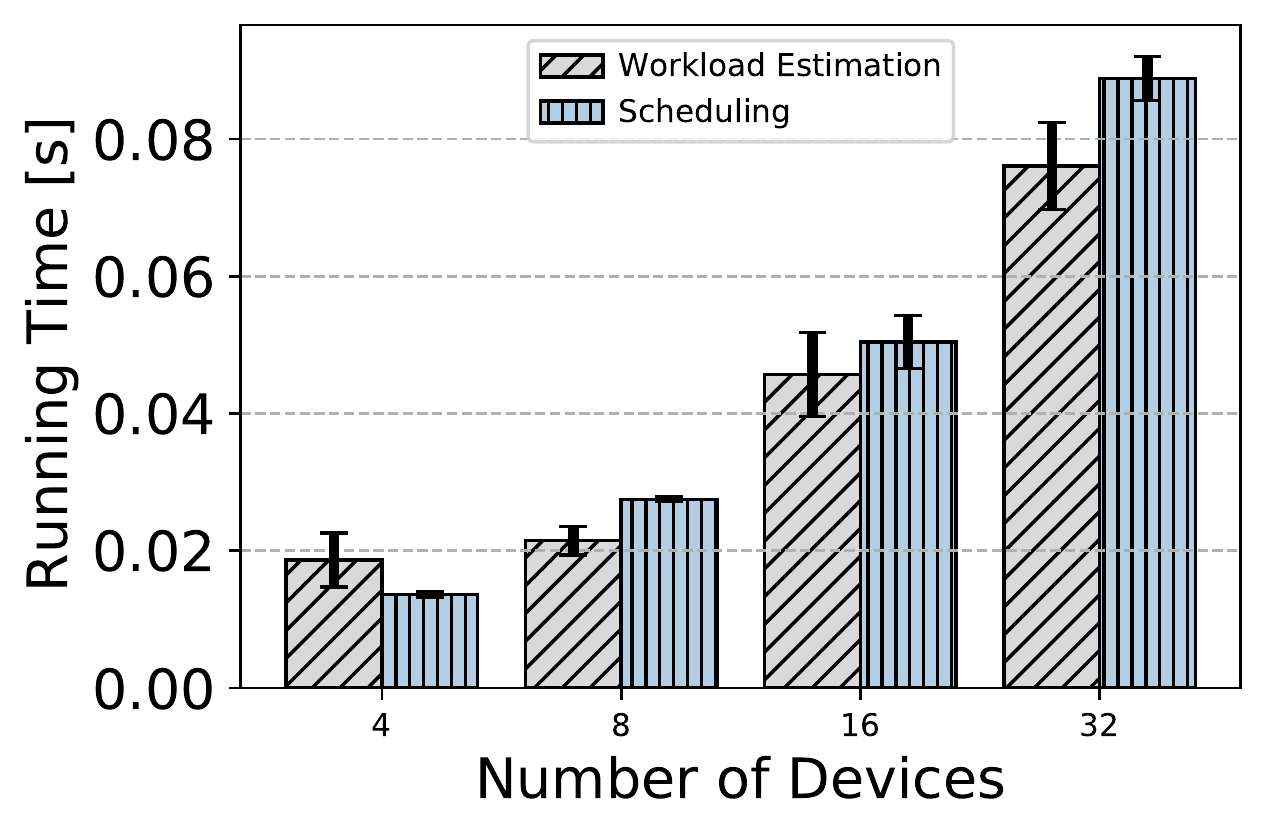}}
     \subfigure[ImageNet with $M_p=100$. ]{\includegraphics[width=0.48\linewidth]{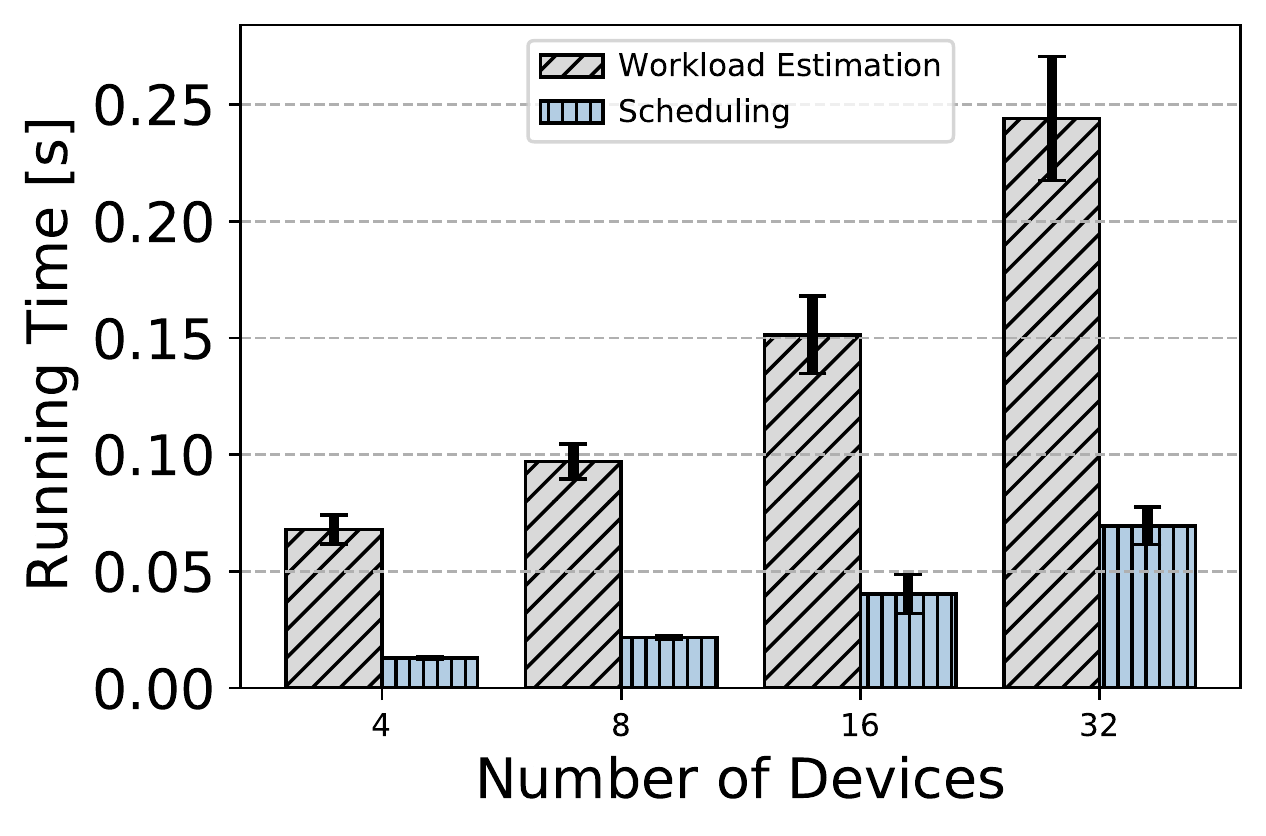}}
    \caption{Averaged time of workload estimation and scheduling per round with different number of devices.}
    \label{fig:ScheTime_C100_SCALE}
\vspace{-0.3cm}
% \vspace{-0.5cm}
\end{figure}

\begin{figure}[ht!]
    \subfigbottomskip=-1pt
    \subfigcapskip=1pt
  \centering
% \!\!\!\!\!\!\!\!
     \subfigure[FEMNIST with $M_p=100$. ]{\includegraphics[width=0.48\linewidth]{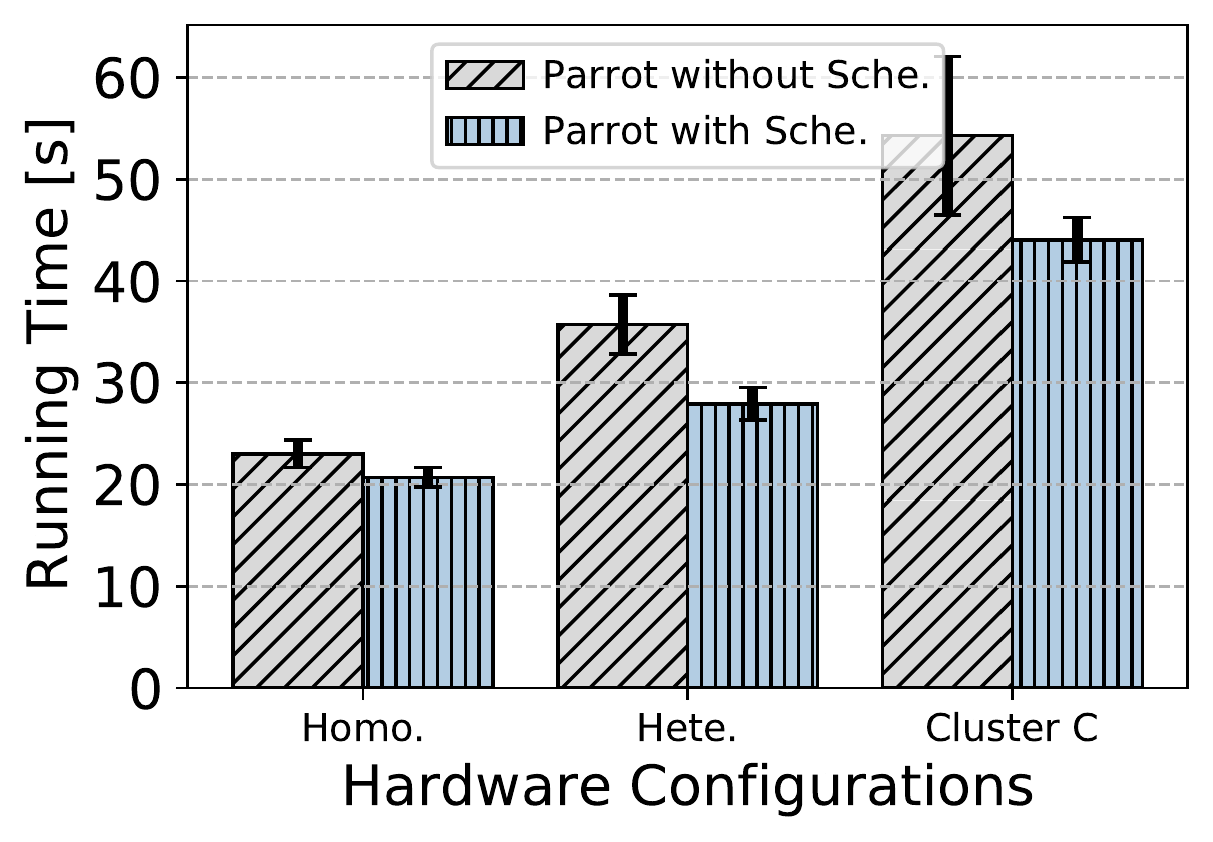}}
     \subfigure[ImageNet with $M_p=100$. ]{\includegraphics[width=0.48\linewidth]{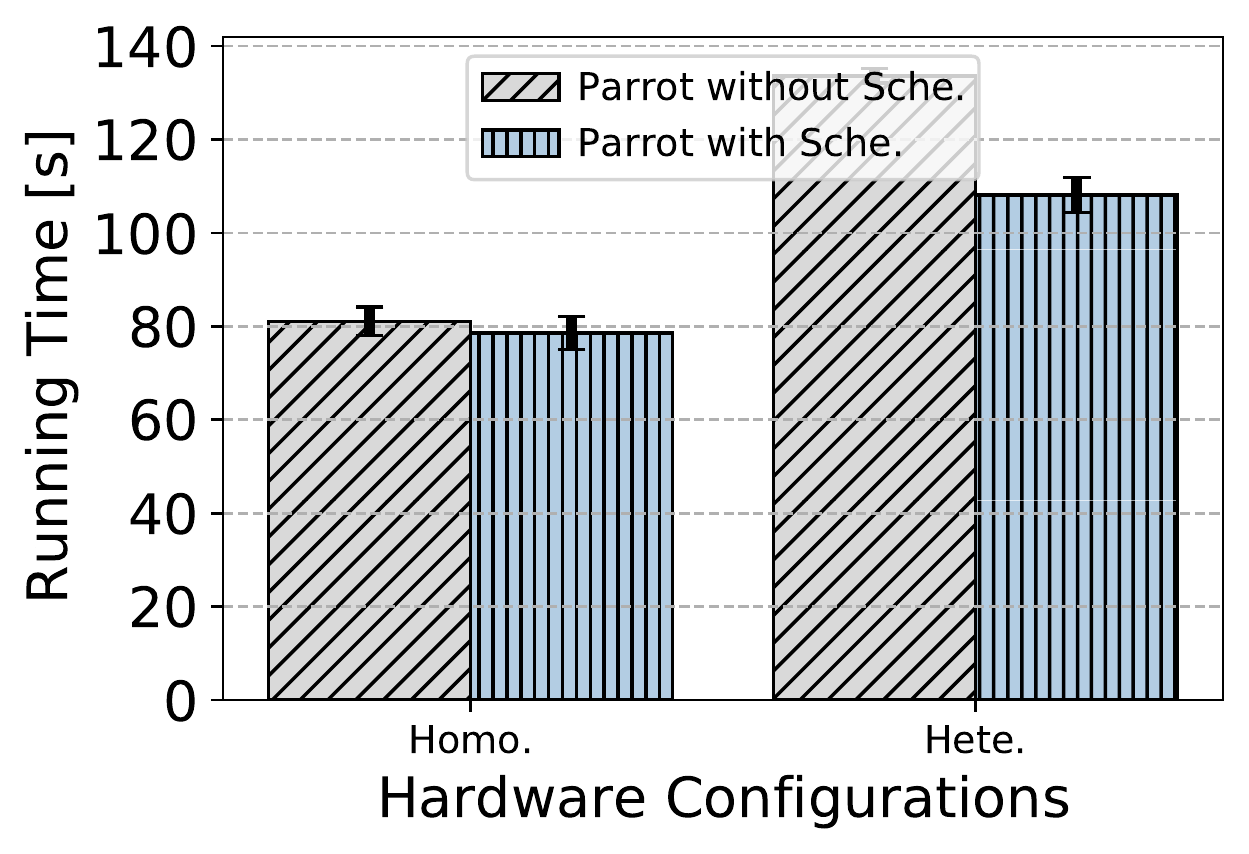}}
    \caption{Running time per round with different hardware configurations.}
    \label{fig:RunTime_C100_HETE_GPU}
\vspace{-0.3cm}
% \vspace{-0.5cm}
\end{figure}

\begin{figure}[ht!]
    \subfigbottomskip=-1pt
    \subfigcapskip=1pt
  \centering
% \!\!\!\!\!\!\!\!
     \subfigure[FEMNIST. ]{\includegraphics[width=0.48\linewidth]{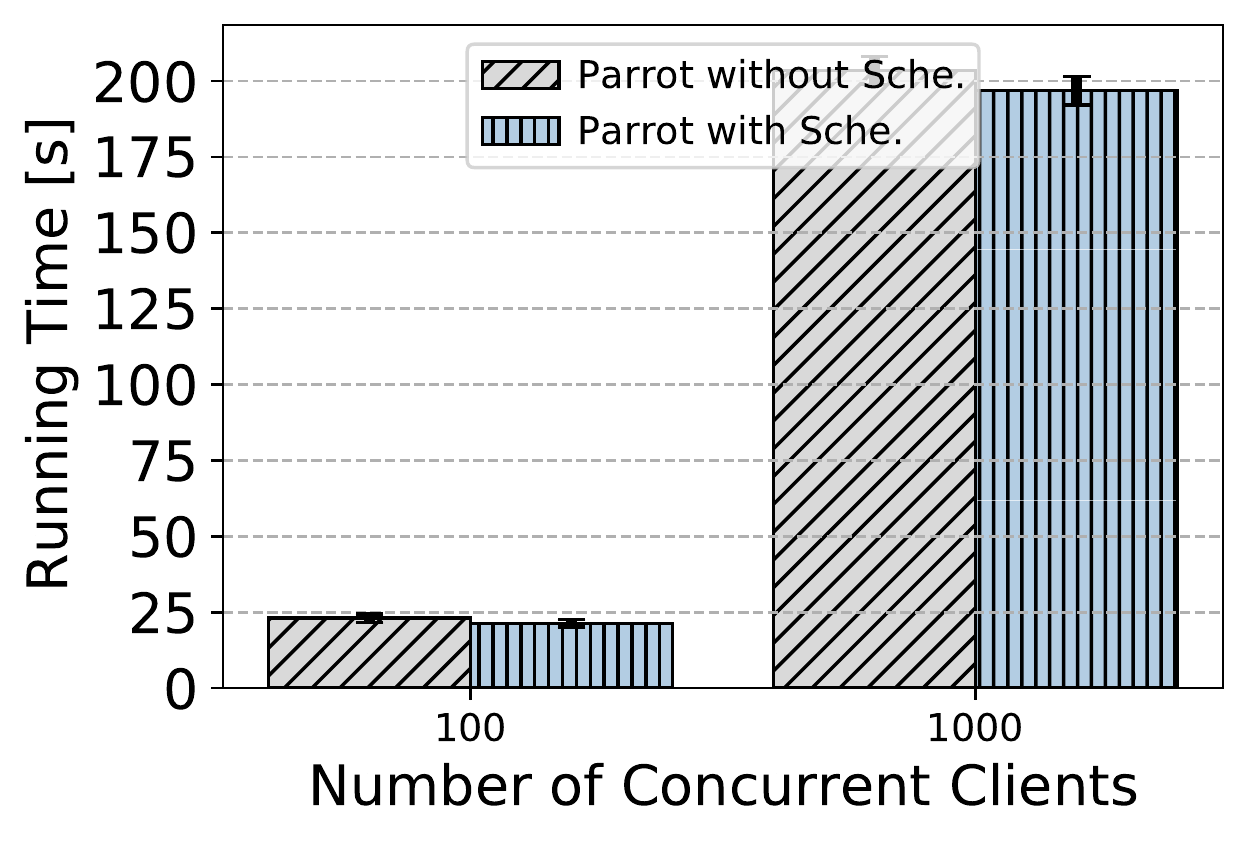}}
     \subfigure[ImageNet. ]{\includegraphics[width=0.48\linewidth]{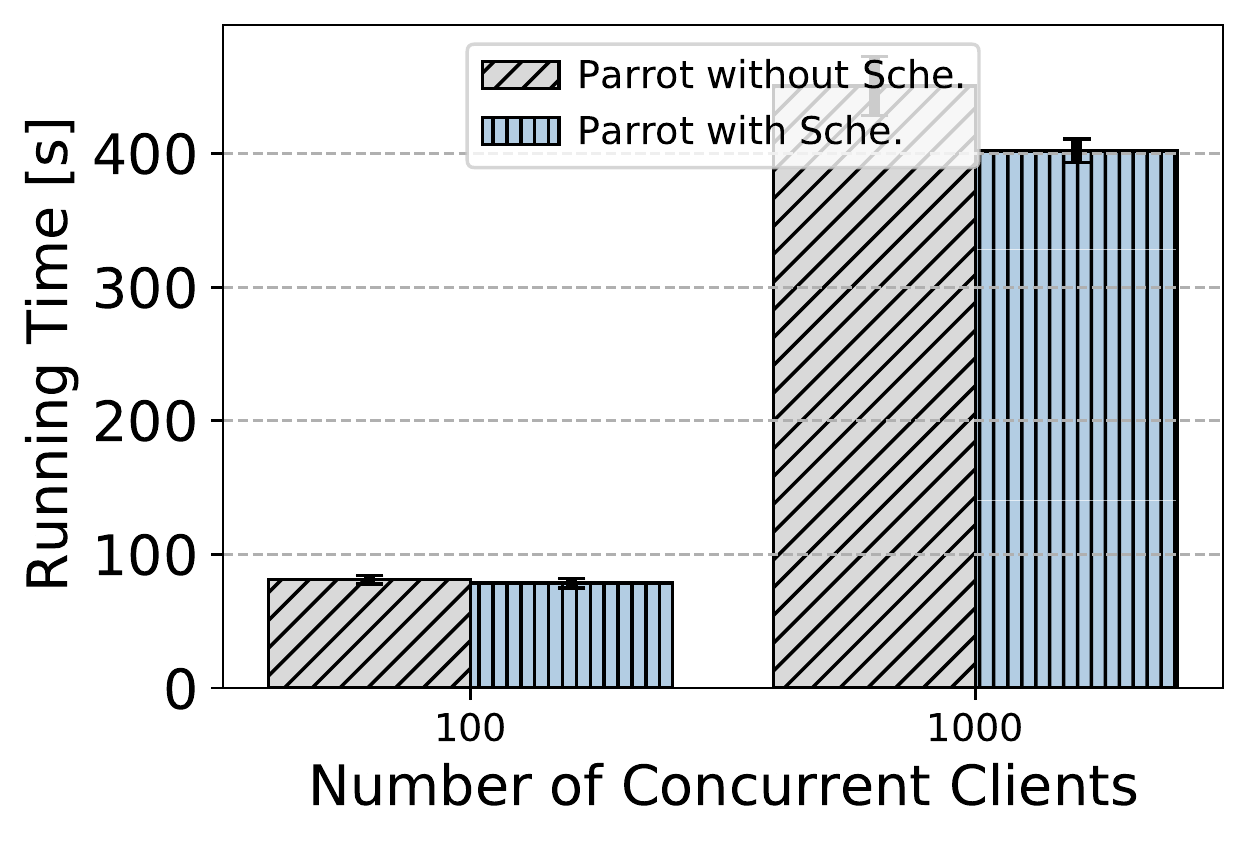}}
    \caption{Running time per round with different number of concurrent clients.}
    \label{fig:RunTime_DifConcurrentClients}
\vspace{-0.3cm}
% \vspace{-0.5cm}
\end{figure}

\begin{figure}[ht!]
    \subfigbottomskip=-1pt
    \subfigcapskip=1pt
  \centering
% \!\!\!\!\!\!\!\!
     \subfigure[Estimation error. ]{\includegraphics[width=0.48\linewidth]{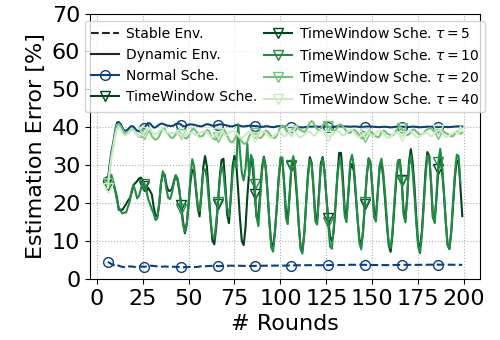}}
     \subfigure[Running time per round. ]{\includegraphics[width=0.48\linewidth]{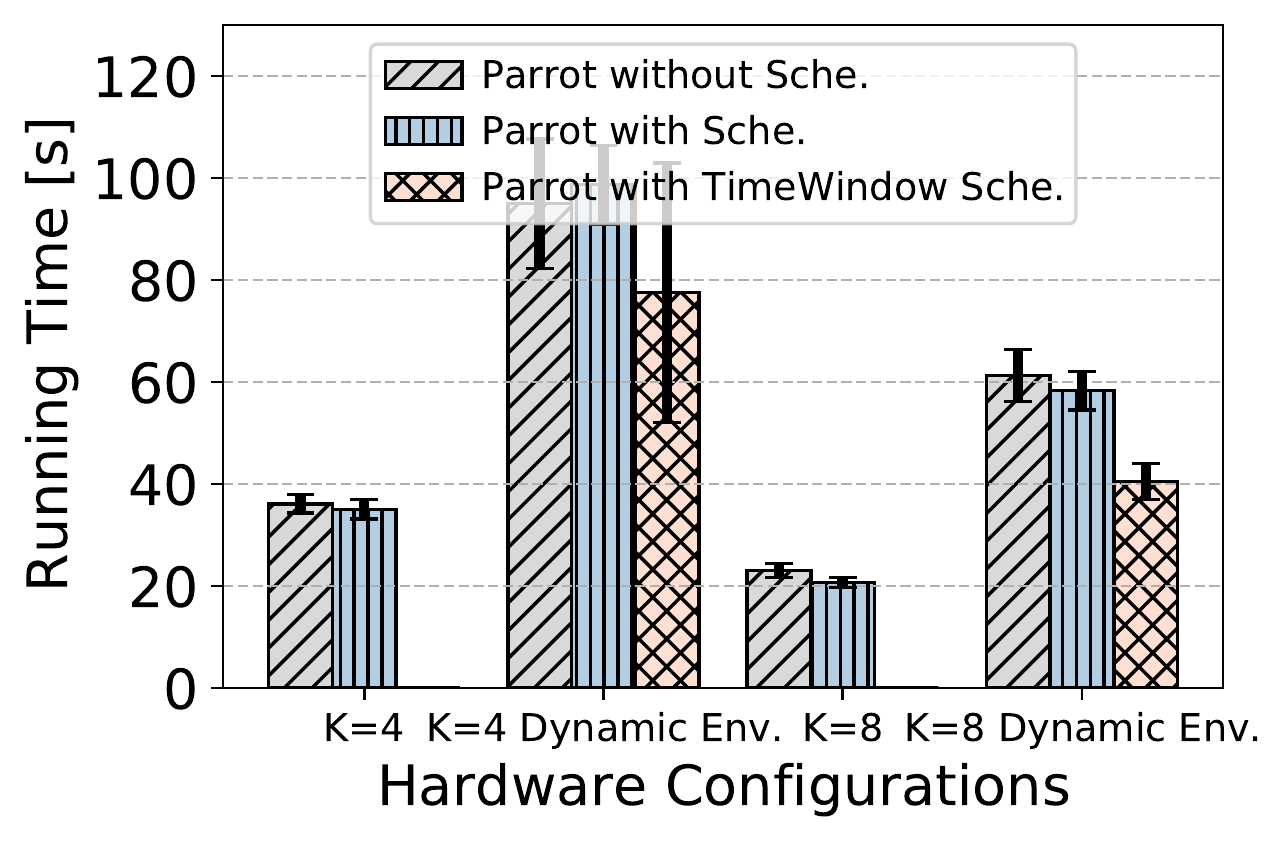}}
    \caption{Running time per round and estimation error of different environments and scheduling algorithms.}
    \label{fig:RunTime_C100_HETE_ENV}
% \vspace{-0.5cm}
\end{figure}

\section{Experiments}\label{sec:exp}

\subsection{Setup}

\textbf{Dataset and models.} To verify that \texttt{Parrot} works well with various datasets and models, we train three different models, including ResNet-18, ResNet-50~\citep{he2016deep} and Albert-V2~\citep{Lan2020ALBERT}) on FEMNIST~\citep{caldas2018leaf}, ImageNet~\citep{russakovsky2015imagenet}, and Reddit~\citep{fedscale}. The detailed information of datasets, models, and hyper-parameters are listed in Appendix.

\textbf{FL settings.} To simulate different FL scenarios, we conduct experiments with the $100$ and $1000$ concurrent clients. And we also try different FL partition methods on ImageNet to see the system performance change of \texttt{Parrot} with different quantity heterogeneity\footnote{Note that only quantity skew~\citep{kairouz2019advances} influences the system performance}.

\textbf{Hardware configuration.} We conduct experiments on three different computer clusters A, B and C, to validate that \texttt{Parrot} performs well in different hardware environments. The detailed hardware and software environments are listed in in Appendix. And we adjust the number of devices ($4 \sim 32$) to see the how much speedup \texttt{Parrot} can gain. We also simulate the heterogeneous-GPU environment (Hete. GPU) and the unstable computing environment (Dyn. GPU) to evaluate that the scheduling is robust to heterogeneous and unstable devices. The simulation methods of heterogeneous-GPUs environments and unstable devices are introduced in Appendix. Note that cluster C itself is a heterogeneous computing cluster, which is used to verify the real-world device heterogeneity rather than simulation.

\textbf{FL algorithms.} To examine that \texttt{Parrot} cant support various FL algorithms, we choose $6$ classic and advanced FL algorithms to simulate. In these algorithms, FedAvg~\citep{mcmahan2017communication} and FedProx~\citep{fedprox} only need to communicate machine learning model parameters. Besides mode parameters, FedNova~\citep{wang2020tackling} needs to communication an aggregation weight, Mime~\citep{mime} needs to communicate local-batch gradient and server optimizer state, SCAFFOLD~\citep{karimireddy2019scaffold} needs to communication control variates. Moreover, Feddyn~\citep{acar2021federated} needs to store local gradients on clients, SCAFFOLD~\citep{karimireddy2019scaffold} needs to store local control variates on clients. We will show that \texttt{Parrot} does not change the performance of these algorithms, and offers friendly supports to them for large-scale simulations.

\subsection{Experiment results}
% \subsection{Comparing Different Frameworks}

\textbf{Comparing Different Frameworks}
We compare \texttt{Parrot} with different FL simulation frameworks, including 
FedML~\citep{chaoyanghe2020fedml}, FedScale~\citep{fedscale} and Flower~\citep{beutel2020flower} to demonstrate the performance efficiency of \texttt{Parrot}. Figure~\ref{fig:EstimationError_C100_Framework} shows the running time of different frameworks in different clusters. \texttt{Parrot} shows the around 1.2 $\sim$ 10 $\times$ acceleration than FedScale, Flower and FedML in all cases. Note that simulation scheme \texttt{SD. Dist} of FedML fails to deploy $M_p=100$ FEMNIST FL on to 4 and 8 devices in cluster A, due to the large memory requirements of \texttt{SD. Dist}.

% \subsection{Implementing FL Algorithms}

\textbf{Implementing FL Algorithms}
We implement various FL algorithms through FedML and \texttt{Parrot}. Note that for SCAFFOLD and FedDyn that need stateful clients, we add client state manager into FedML to implement them. The convergence results of different algortihms are shown in Figure~\ref{fig:TestAccFMNISTAlgo}, showing that algorithms implemented by \texttt{Parrot} can perform similar or higher test accuracy than the \texttt{SD Dist.} scheme. Figure~\ref{fig:TestAccFMNISTspe_param_algo} shows that \texttt{Parrot} can support algorithms that need to communication special parameters well. Figure~\ref{fig:TestAccFMNISTstateful_algo} shows \texttt{Parrot} can implement stateful-clients algorithms with large-scale experiments.

The averaged running times of different FL algorithms are shown in Figure~\ref{fig:RunTimeFMNISTAlgo}. It demonstrates that scheduling of \texttt{Parrot} can help accelerate running all chosen FL algorithms. And some algorithms that needs extra calculation overheads show more running time reduction.

\vspace{-0.1cm}
\begin{table}[ht!]
\vspace{-0.1cm}
\centering
\caption{GPU Memory Costs of Different FL Tasks.} 
\vspace{0pt}
% \small{
\footnotesize{
\begin{tabular}{lcc|ccc}
\toprule[1.5pt]
&  &  & \multicolumn{3}{c}{GPU Memory Cost (MB)} \\
\midrule[0.5pt]
Dataset & $M_p$       & $K$    & SP   & SD Dist. & \makecell[c]{FA Dist. \& \\  Parrot} \\
\midrule[0.5pt]
FEMNIST & $100$   & 8  & 1,122  & 112,200          & 8,976   \\
FEMNIST & $100$   & 16 & 1,122  & 112,200          & 17,952   \\
ImageNet & $1000$ & 8  & 3,305   & 3305,000   & 26,440   \\
ImageNet & $1000$  & 16 & 3,305  & 3305,000   & 52,880   \\
\bottomrule[1.5pt] 
\end{tabular}
}
% \vspace{-0.5cm}
\label{tab:GPUmemoryCost}
\end{table}

% \begin{figure}[htb!] 
% \small
%   \setlength{\abovecaptionskip}{0.cm}
%   \centering
%     \includegraphics[width=1.0\linewidth]{results/FEMNIST_frameworks.pdf} 
%     \caption{FEMNIST with different frameworks.} 
%     \label{fig:FEMNIST_frameworks} 
% \end{figure}
% \vspace{-0.1cm}

% \begin{figure}[htb!] 
% \small
%   \setlength{\abovecaptionskip}{0.cm}
%   \centering
%     \includegraphics[width=1.0\linewidth]{results/Reddit_frameworks.pdf} 
%     \caption{Reddit.} 
%     \label{fig:Reddit_frameworks} 
% \end{figure}
% \vspace{-0.1cm}

% \subsection{Workload Estimation}

% Running times of simulating clients on devi
% Figure~\ref{fig:EstimationHomoEnv} and~\ref{fig:EstimationHeteEnv} show the effect of workload estimation of different FL tasks in different hardware environments. 

% Figure~\ref{fig:EstimationHomoEnv}(a)

% \subsection{Different Number of Devices}

\textbf{Different Number of Devices}
We vary the number of devices to validate the scalability of \texttt{Parrot} as shown in Figure~\ref{fig:RunTimeRound_C100_SCALE}. The results show that \texttt{Parrot} scales well with more clients. And the times of workload estimation and scheduling shown in Figure~\ref{fig:ScheTime_C100_SCALE} increase almost linearly with the number of devices. Comparing with the running time, the scheduling consumes little time.

% Figure~\ref{fig:EstimationError_C100_SCALE} shows the error of workload estimation decreases wi

% \subsection{Different Number of Concurrent Clients}
\textbf{Different Number of Concurrent Clients}
Figure~\ref{fig:RunTime_DifConcurrentClients} shows the results of simulating FL with different number of concurrent clients, i.e. 100 and 1000. Our scheduling shows benefits for both different scales.

% \subsection{GPU Memory Cost}

\textbf{GPU Memory Cost}
Table~\ref{tab:GPUmemoryCost} shows the GPU memory costs of simulating different FL experiments. It shows that the \texttt{SD Dist.} needs massive memory to simulate large-scale experiments with $M_p = 1000$. And it also needs large memory to simulate experiments with  $M_p = 100$.

% \subsection{Heterogeneous Computing Devices}

\textbf{Heterogeneous Computing Devices}
Figure~\ref{fig:EstimationHomoEnv} shows that \texttt{Parrot} can accurately estimate the running times of tasks on both homogeneous or heterogeneous devices. Thus, as shown in Figure~\ref{fig:RunTime_C100_HETE_GPU}, the running times of \texttt{Parrot} with scheduling are much lower than \texttt{Parrot} without scheduling.

% \subsection{Dynamic Environment}

\textbf{Dynamic Environment}
In the dynamic environments, due to the unstable computing performance of devices, the running times of the save tasks may continuously change. Thus, it is hard to utilize all historical data to fit our workload model~\ref{eq:TimeModelDifDevice}. As shown in Figure~\ref{fig:RunTime_C100_HETE_ENV}(a), scheduling with all historical data fails to estimate the workload in the dynamic environments. Thus the running time of it is similar with \texttt{Parrot} without scheduling as shown in Figure~\ref{fig:RunTime_C100_HETE_ENV}(b). The Time Window scheduling can estimate the workload more accurately, thus achieving the reduction of running time. 
% But Figure~\ref{fig:RunTime_C100_HETE_ENV}(a) also shows that different $\tau$ severely influences the performance of estimation. 

% \section{Discussion}\label{sec:discuss}

\section{Conclusion}\label{sec:conc}
In this work, we demystify the challenges and bottlenecks of simulating FL, and design a new FL system named as \texttt{Parrot}. \texttt{Parrot} provides an efficient FL simulator and drastically reduces the hardware requirements. With the state manager, \texttt{Parrot} provides friendly support to a wide range of FL algorithms with stateful clients. The scheduling component of \texttt{Parrot} highly accelerates simulating FL. With the general and friendly APIs, users can seamlessly migrate the verified FL algorithms and models from the simulation into real-world deployment effortlessly without changing their code.

% The current design of \texttt{Parrot} has not considered the 

% \section*{Acknowledgements}

\nocite{langley00}

\bibliography{cite}
\bibliographystyle{mlsys2022}

%%%%%%%%%%%%%%%%%%%%%%%%%%%%%%%%%%%%%%%%%%%%%%%%%%%%%%%%%%%%%%%%%%%%%%%%%%%%%%%
%%%%%%%%%%%%%%%%%%%%%%%%%%%%%%%%%%%%%%%%%%%%%%%%%%%%%%%%%%%%%%%%%%%%%%%%%%%%%%%
% SUPPLEMENTAL CONTENT AS APPENDIX AFTER REFERENCES
%%%%%%%%%%%%%%%%%%%%%%%%%%%%%%%%%%%%%%%%%%%%%%%%%%%%%%%%%%%%%%%%%%%%%%%%%%%%%%%
%%%%%%%%%%%%%%%%%%%%%%%%%%%%%%%%%%%%%%%%%%%%%%%%%%%%%%%%%%%%%%%%%%%%%%%%%%%%%%%
\newpage
\appendix
\onecolumn

\section*{Appendix}

\section{Experiment Configuration}\label{sec:apx_exp}
In this section, we show the detailed experiment configurations and more results.

\begin{table*}[ht!]
\centering
\caption{Details of datasets and hyper-parameters.} 
\vspace{1pt}
% \small{
\resizebox{\columnwidth}{!}{
\begin{tabular}{lcccccccc}
\toprule[1.5pt]
Dataset &  Partition & \# Clients & $M_p$ &   Model    & Model Params.  &  learning rate &  batch size & local epochs \\
\midrule[1.5pt]
FEMNIST & Natural & 3,400  & $100$   & ResNet-18  & 11M  &  0.05 & 20 &  10  \\
FEMNIST & Natural &  3,400  & $1000$   & ResNet-18 & 11M  &  0.05 & 20 &  10  \\
ImageNet(a) & Dirichlet(0.1) & 10,000 &  $100$ & ResNet-50  & 23M   &  0.05 & 20 &  2  \\
ImageNet(a) & Dirichlet(0.1) & 10,000  & $1000$  & ResNet-50 & 23M  &  0.05  & 20 & 1  \\
ImageNet(b) & Quantity Skew(5.0) & 10,000 &  $100$ & ResNet-50  & 23M   &  0.05 & 20 &  2  \\
Reddit & Natural & 1,660,820 & $100$ &  Albert-Base-v2 & 11M & 0.0005  & 20  & 5   \\
\bottomrule[1.5pt] 
\end{tabular}
}
\vspace{-0.1cm}
\label{tab:hyperparameter}
\end{table*}

\begin{table*}[ht!]
\centering
\caption{Hardware Configurations.} 
\vspace{1pt}
% \small{
\resizebox{\columnwidth}{!}{
\begin{tabular}{lcccccc}
\toprule[1.5pt]
Name &  CPU & GPU  & Network Bandwidth & Operation System & CUDA &  PyTorch \\
\midrule[1.5pt]
Cluster A &  \makecell[c]{Intel(R) Xeon(R) \\Gold 5115 CPU @ 2.40GHz} & \makecell[c]{4$\times$ RTX 2080 Ti \\on 8 nodes}   &  \makecell[c]{10Gbps or\\ 100Gbps} &  Ubuntu 18.04.6  & V11.3  & V1.12.1 \\
\midrule[0.5pt]
Cluster B &  \makecell[c]{Intel(R) Xeon(R) \\Gold 5220R CPU @ 2.20GHz} & 8 $\times$ Quadro RTX 5000  &  10Gbps   & Ubuntu 18.04.5  & V11.2  & V1.12.1 \\
\midrule[0.5pt]
Cluster C & \makecell[c]{Intel(R) Xeon(R) \\CPU E5-2630 v3 @ 2.40GHz}  & \makecell[c]{node1: 4$\times$Tesla K80,\\ node2: 2$\times$Tesla P40, \\ node3: 2$\times $Tesla P40}    & \makecell[c]{10Gbps or \\ 100Gbps}    &  \makecell[c]{Oracle Linux \\Server 8.6} & V11.4  &  V1.10.1 \\
\bottomrule[1.5pt] 
\end{tabular}
}
\label{tab:hardware}
\end{table*}

\textbf{Simulation of heterogenous-GPUs envirionments.}
To test the adaptability of \texttt{Parrot} in different hardware environments. We simulate the heterogenous-GPUs envirionments (Hete. GPU) and unstable devices (Dyn. GPU). 
Because heterogeneous GPUs show different computing time on the same tasks, we pre-assign different heterogeneous ratio $\left\{ \eta_k | k \in \mathcal{K} \right\}$ to devices. After each local training on task $m$ with time $\hat{T}^r_{m,k}$, devices $k$ will sleep for a period of time $ \eta_k\hat{T}^r_{m,k}$. Thus, the homogeneous GPUs will show different computing times. Lower $\eta_k$ means more powerful GPUs. Now, the server will obtain the running time $ \eta_k\hat{T}^r_{m,k}$ to conduct workload estimation instead of $\hat{T}^r_{m,k}$.

\textbf{Simulation of unstable devices.}
Similar with simulating heterogenous-GPUs envirionments, we exploit different sleep time to simulate unstable devices. Specifically, we generate the ratio of sleep time at different communication rounds with different clients, i.e. $(1 + cos(3.14r/R + k))$. Thus, during the total training process, devices have different computing performance with different rounds.

% \begin{figure}[ht!]
%     % \setlength{\abovedisplayskip}{-2pt}
%     % \setlength{\abovecaptionskip}{-2pt}
%     \subfigbottomskip=-1pt
%     \subfigcapskip=1pt
%   \centering
% % \!\!\!\!\!\!\!\!
%      \subfigure[FEMNIST with 1000 concurrent clients. ]{\includegraphics[width=0.24\linewidth]{runtime_bar/RunTimeRound_e_fem_c1000_scal.pdf}}
%      \subfigure[ImageNet with 1000 concurrent clients. ]{\includegraphics[width=0.24\linewidth]{runtime_bar/RunTimeRound_e_img_c1000_scal.pdf}}
%     \caption{Averaged running time of 1000 concurrent clients.}
%     \label{fig:RunTime_C1000_SCALE}
% \vspace{-0.3cm}
% % \vspace{-0.5cm}
% \end{figure}

% \section{Examples of Implementing FL Algorithms}

% \begin{lstlisting}[language=Python, caption=Python example]
% import numpy as np
    
% def incmatrix(genl1,genl2):
%     m = len(genl1)
%     n = len(genl2)
%     M = None #to become the incidence matrix
%     VT = np.zeros((n*m,1), int)  #dummy variable
    
%     #compute the bitwise xor matrix
%     M1 = bitxormatrix(genl1)
%     M2 = np.triu(bitxormatrix(genl2),1) 

%     for i in range(m-1):
%         for j in range(i+1, m):
%             [r,c] = np.where(M2 == M1[i,j])
%             for k in range(len(r)):
%                 VT[(i)*n + r[k]] = 1;
%                 VT[(i)*n + c[k]] = 1;
%                 VT[(j)*n + r[k]] = 1;
%                 VT[(j)*n + c[k]] = 1;
                
%                 if M is None:
%                     M = np.copy(VT)
%                 else:
%                     M = np.concatenate((M, VT), 1)
                
%                 VT = np.zeros((n*m,1), int)
    
%     return M
% \end{lstlisting}

\end{document}